\documentclass[conference]{IEEEtran}
\IEEEoverridecommandlockouts
\usepackage{microtype}
\usepackage{graphicx}
\usepackage{booktabs} % for professional tables

\usepackage{hyperref}

\usepackage{multicol}
\usepackage{amsmath,amssymb,amsfonts}

\usepackage{algorithmic}
\usepackage{graphicx}
\usepackage{textcomp}
\usepackage{xcolor}
\usepackage{caption}
\newcommand{\ignore}[1]{}
\usepackage[normalem]{ulem}
\usepackage{fancyhdr}
\usepackage[normalem]{ulem}
\usepackage{makecell}
\usepackage{svg}
\usepackage{xcolor}
\usepackage{xspace}
\usepackage{bbold}
\usepackage{pbox}
\usepackage[rightcaption]{sidecap}
\sidecaptionvpos{figure}{c}
\usepackage{physics}
\usepackage{soul}
\usepackage{enumitem}
\usepackage{comment}
\usepackage{tikz}
\usepackage{adjustbox}
\usepackage{bibunits}
\usepackage{subfig}
\usepackage{mathtools}
\usepackage{float}
\definecolor{Awesome}{rgb}{1.0, 0.08, 0.58}

\definecolor{fabulous}{rgb}{1.0, 0.0, 0.5}

\newcommand{\MP}{\textit{Quantum Mantissa}\xspace}
\newcommand{\BP}{\textit{Quantum Integer}\xspace}
\newcommand{\QE}{\textit{Quantum Exponent}\xspace}
\newcommand{\QEshort}{\textit{QE}\xspace}
\newcommand{\QEGshort}{\textit{QE+G}\xspace}
\newcommand{\QMshort}{\textit{QM}\xspace}

\newcommand{\QMQEGshort}{\textit{QM+QE+G}\xspace}
\newcommand{\QMQEshort}{\textit{QM+QE}\xspace}
\newcommand{\QIshort}{\textit{QI}\xspace}
\newcommand{\Gshort}{\textit{+G}\xspace}

\newcommand{\BWMshort}{\textit{BWM}\xspace}
\newcommand{\BWEshort}{\textit{BWE}\xspace}
\newcommand{\BWEGshort}{\textit{BWE+G}\xspace}
\newcommand{\BWMBWEshort}{\textit{BWM+BWE}\xspace}

\newcommand{\MC}{\textit{BitWave}\xspace}
\newcommand{\EC}{\textit{Gecko}\xspace}
\newcommand{\DPRL}{\textit{Schr\"{o}dinger's FP}\xspace}

\newcommand{\DPRTITLE}{Schr\"{o}dinger's FP\xspace}

\newcommand{\GMC}{\textit{$\mathit{\mathrm SFP}_\mathit{\mathrm BW}$}\xspace}
\newcommand{\GMP}{\textit{$\mathit{\mathrm SFP}_\mathit{\mathrm Q}$}\xspace}
\newcommand{\GMCG}{\textit{$\mathit{\mathrm SFP}_\mathit{\mathrm BW+G}$}\xspace}
\newcommand{\GMPG}{\textit{$\mathit{\mathrm SFP}_\mathit{\mathrm Q+G}$}\xspace}

\DeclarePairedDelimiter\floor{\lfloor}{\rfloor}

\begin{document}
\title{\DPRTITLE: Training Neural Networks with Dynamic Floating-Point Containers\thanks{}}

\author{
\begin{tabular}{cccc}
\begin{minipage}[t]{0.2\textwidth}
\centering
Milo\v{s} Nikoli\'{c}\textsuperscript{1}\\
\textit{University of Toronto} \\
Toronto, Canada
\end{minipage} &
\begin{minipage}[t]{0.21\textwidth}
\centering
Enrique Torres Sanchez\textsuperscript{1} \\
\textit{University of Toronto} \\
Toronto, Canada
\end{minipage} &

\begin{minipage}[t]{0.2\textwidth}
\centering
Jiahui Wang\textsuperscript{2} \\
\textit{Qualcomm} \\
Toronto, Canada
\end{minipage} &
\begin{minipage}[t]{0.2\textwidth}
\centering
Ali Hadi Zadeh\textsuperscript{2} \\
\textit{1QBit} \\
Toronto, Canada
\end{minipage} \\ \\

\begin{minipage}[t]{0.2\textwidth}
\centering
Mostafa Mahmoud\textsuperscript{2} \\
\textit{AMD} \\
Toronto, Canada
\end{minipage} &
\begin{minipage}[t]{0.2\textwidth}
\centering
Ameer Abdelhadi\textsuperscript{2} \\
\textit{McMaster University} \\
Toronto, Canada
\end{minipage} &

\begin{minipage}[t]{0.2\textwidth}
\centering
Kareem Ibrahim \\
\textit{University of Toronto} \\
Toronto, Canada
\end{minipage} &
\begin{minipage}[t]{0.2\textwidth}
\centering
Andreas Moshovos \\
\textit{University of Toronto} \\
Toronto, Canada
\end{minipage}
\end{tabular}
}

\maketitle

\begin{abstract}
The transfer of tensors from/to memory during neural network \emph{training} dominates time and energy. To improve energy efficiency and performance, research has been exploring ways to use narrower data representations. So far, these attempts relied on user-directed trial-and-error to achieve convergence. We present methods that relieve users from this responsibility. Our methods dynamically adjust the size and format of the floating-point containers used for activations and weights during training, achieving adaptivity across three dimensions: i)~which datatype to use, ii)~on which tensor, and iii)~how it changes over time. The different meanings and distributions of exponent and mantissas lead us to tailored approaches for each. We present two lossy \textit{pairs} of methods to eliminate as many mantissa and exponent bits as possible without affecting accuracy. \emph{\MP} and \emph{\QE} are machine learning compression methods that tap into the gradient descent algorithm to \emph{learn} the minimal mantissa and exponent bitlengths on a per-layer granularity. They automatically \textit{learn} that many tensors can use just 1 or 2 mantissa bits and 3 or 4 exponent bits. Overall, the two machine learning methods reduce the footprint by $4.74\times$. Alternatively, \emph{\MC} observes changes in the loss function during training to adjust mantissa and exponent bitlengths network-wide, yielding a $3.19\times$ reduction in footprint. Finally, we present an optional method, \EC, to exploit the naturally emerging, lop-sided exponent distribution to \textit{losslessly} compress resulting exponents from \QE or \MC and, on average, improve compression rates to $5.64\times$ and $4.56\times$.
\end{abstract}

\section{Introduction}

\footnotetext[1]{Correspondence to: Milo\v{s} Nikoli\'{c} \
\textless{}milos.nikolic@mail.utoronto.ca\textgreater{} and Enrique Torres Sanchez \textless{}enrique.torres@mail.utoronto.ca\textgreater{}.}

\footnotetext[2]{All work completed at the University of Toronto.}

While training neural networks is both computationally and data demanding, it is the memory transfers to off-chip DRAM for \textit{stashing} (i.e., saving and much later recovering) activation and weight tensors that dominate execution time and energy~\cite{GIST}. The per batch data volume easily surpasses on-chip memory capacities, necessitating off-chip DRAM accesses which are up to two orders of magnitude slower and more energy expensive~\cite{HorowitzEnergy}. Reducing this overhead has been receiving attention throughout the software/hardware stack and is also our goal. 

The most direct way to reduce tensor volume is by using datatypes that use fewer bits per value, e.g.,  BFloat16~\cite{bfloat16},  half-precision floating-point (FP16), dynamic floating-point~\cite{DBLP:conf/iclr/0002MMKAB0VKGHD18}, flexpoint~\cite{Koster:2017:FAN:3294771.3294937}), or fixed-point~\cite{DBLP:conf/iclr/0002MMKAB0VKGHD18,mixedP,nvidia_mixedP,Drumond:2018:TDH:3326943.3326985}). This reduces memory traffic and footprint, improving energy efficiency and execution time. In the past, training typically used single precision 32b floating-point (FP32), as it was believed to yield the best accuracy. However, recent research has shown that using more compact datatypes can still achieve good results while reducing memory usage. Some works have even pushed the limits of datatype efficiency by using 8b~\cite{IBM_8bit} and 4b~\cite{IBM_4bit} datatypes in \textit{certain} cases. Industry is even exploring the use of 8b floating point with different mantissa/exponent ratios to meet the specific needs of tensors~\cite{fp8_Nvidia_Intel_Arm} and even shorter formats~\cite{rouhani2023microscaling}. As industry is expanding support for leaner datatypes the following challenges remain:
\begin{itemize}[noitemsep,nolistsep,topsep=0pt,parsep=0pt,partopsep=0pt, leftmargin=7.5mm]
\item To achieve convergence current approaches rely \textit{exclusively} on trial-and-error: It is up to the user to carefully select which datatype to use for each tensor. This often necessitates changes to the training recipe and the inclusion of additional operations such as loss scaling~\cite{nvidia_mixedP}. Convergence is not guaranteed and can be evaluated only \textit{post mortem}.
\item Universally, all methods store weights in full-precision as the backward pass performs minuscule updates that cannot be represented with the leaner datatype.
\item The datatypes are statically chosen offering no opportunity to amend the choice if accuracy suffers (e.g., significant drop with deeper networks identified by IBM~\cite{IBM_4bit}). 
\item Even where successful, these methods still use a scant repertoire of bitlengths (e.g., tensors fitting in 5b have to use 8b, a nearly 2x increase), leaving a lot of opportunity for memory overhead reduction untapped. 
\item They require hardware changes to allow computation with the leaner datatypes.
\end{itemize}

This work automates and fuses \textit{into training itself} the process of datatype discovery improving execution time and energy efficiency. Given that floating-point remains the datatype of choice to ensure convergence, we focus on \textit{automatic} floating-point datatype selection with the goal being to reduce memory traffic during training. Our methods:
\begin{itemize}[noitemsep,nolistsep,topsep=0pt,parsep=0pt,partopsep=0pt, leftmargin=7.5mm]
\item \textit{Dynamically} and \textit{continuously} adjust the \textit{mantissa} and the \textit{exponent bitlengths} for floating-point activations and/or weights for stashed tensors, and do so \textit{transparently} at no additional burden to the user.
\item Are adaptable across three dimensions: The first two automate what is currently done by hand: \textit{which} datatype to use for \textit{which} tensor. Uniquely, our methods adapt these datatypes over \textit{time}.  
\item Adapt the exponent bitlengths to their actual content using only as many bits as necessary to store their value. Most exponents end up using a lot fewer bits than statically selected datatypes.
\item Store values in memory with only as many bits as necessary while expanding values to the closest available datatype supported by the accelerator.
\item In addition to accelerating training, our methods can inform efforts for selecting more efficient datatypes for inference such as that by~\cite{fp8_Nvidia_Intel_Arm}, \cite{ rouhani2023microscaling} or~\cite{IBM_4bit}. 
\item As a by-product, quantize the networks to efficient datatypes which benefits inference.
\end{itemize}
 
Our solution is \DPRL, a \textit{family} of two methods that learn exponents and mantissa bitlengths, and an \textit{optional} lossless exponent compression method \textit{\EC}:

\noindent\textbf{Quantum Mantissa \& Exponent:}  The first method comprises \MP (\QMshort) and \QE (\QEshort), and harnesses the training algorithm itself to \textit{learn} on-the-fly the per tensor mantissa and exponent bitlengths which it continuously adapts per batch. \QMshort and \QEshort introduce a learning parameter per tensor and a regularizer that include the effects of the mantissa and exponent bitlengths. Learning the bitlengths incurs a negligible overhead compared to the resulting reduction in off-chip traffic. Experiments show that: 1)~they reduce bitlengths considerably, more so for mantissas, 2)~the bitlengths vary per tensor and 3)~fluctuate throughout, capturing benefits that wouldn't be possible with a static network-wide choice of datatype.

\noindent\textbf{\MC: } \MC approaches the training as a black-box observing the effect of adjusting mantissa and exponent bitlengths on its progress. It uses a simple linear regression of a history of losses (observed per-batch) to adjust the mantissa and exponent bitlengths for the whole network. As long as the network seems to be improving, \MC will attempt to shorten them; otherwise, it will increase them. \MC proves effective, albeit with lower bitlength reductions compared to \QMQEshort, since: 1)~they harness the training process to learn the optimal bitlengths, and 2)~they adjust bitlengths per layer whereas \MC does so network-wide to reduce the search space.

\noindent\textbf{\EC:} On top of \QMQEshort and \MC exponent bitlength reduction, we introduce an \textit{optional} method, \EC, which exploits their biased distribution that naturally occurs during training~\cite{FPRaker}. \EC stores exponents using only as many bits as necessary to represent their value, outperforming any statically chosen bitlength. \EC chooses the bitlength per group of values to reduce metadata overhead achieving high encoding efficiency.  Encoding values in DRAM using variable length containers is standard practice in systems for deep learning,~\cite{Deep_Compression,ShapeshifterMICRO,han_eie:isca_2016}. 

\noindent\textbf{Reducing Off-Chip Traffic: } We demonstrate that our methods boost energy efficiency and performance by transparently encoding values as they are being stashed to off-chip DRAM, and decoding them to their original format as they are being read back. To do so, we introduce \textit{(de)compressor} units in front of the memory controller leaving the rest of the on-chip memory hierarchy and compute cores unchanged. Future work can investigate using \DPRL to boost computation throughput as well.

To maximize benefits, we present a hardware-assisted implementation of \DPRL (a software-only implementation is possible as well but is left for future work); the inclusion of specialized hardware units is now commonplace among all hardware vendors. Appendix~\ref{sec:hardware} presents efficient hardware (de)compressors that operate on groups of unmodified floating-point values. The units accept external mantissa and exponent length signals and pack values maintaining DRAM-friendly long, sequential accesses. The decompressors expand such compressed blocks back into the original floating-point format.

\DPRL will generally work in conjunction with methods that partition, distribute, or reschedule the training work to improve energy efficiency and performance, or that can improve accuracy for a preselected datatype.

We highlight the following key contributions and experimental findings from \DPRL:
\begin{itemize}
    \item We introduce two machine learning based methods: \MP (\QMshort) and \QE (\QEshort), which harness the training algorithm itself to dynamically learn per-tensor mantissa and exponent bitlengths, adjusting them continuously with each batch. \QMQEshort reduces the memory footprint by $4.74\times$ on average (range: $3.35\times$ to $13.23\times$). The \QMQEshort experiments demonstrate variability in the mantissa and exponent bitlengths across different tensors, thereby highlighting the superiority of this per-tensor approach.
    \item We introduce two loss observation based methods: \MC \textit{Mantissa} (\BWMshort) and \MC \textit{Exponent} (\BWEshort), which approach the training process as a block-box, and observe the effect of adjusting mantissa and exponent bitlengths, via the loss function. \MC reduces the memory footprint without noticeable loss of accuracy by $3.19\times$ on average (range: $2.24\times$ to $8.91\times$). Crucially, \MC stays transparent to the process and has negligible overhead.
    \item We introduce \EC, a lossless exponent group compression method for training. Our work shows that this method can further boost the \QMQEshort and \MC footprint reduction to  $5.64\times$ on average (range: $3.73\times$ to $17.66\times$) and $4.56\times$ on average (range: $3.07\times$ to $9.74\times$), respectively.
    \item Indicatively, for  an accelerator using BFloat16 and with a peak throughput of 16 TFLOPS, \GMPG and \GMCG improve energy efficiency by $3.07\times$ and $2.71\times$. When the accelerator uses FP8 instead, our aforementioned methods improve energy efficiency by $2.26\times$ and $2.00\times$ respectively. 
\end{itemize}

\label{sec:related_work}

\section{Training with Efficient Datatypes}

The question of which \textit{training} datatype strikes the right balance among accuracy, energy and time remains open.
Recently, we have seen success in training with leaner floating-point such as half-precision FP16 and BFloat16~\cite{bfloat16}. These approaches can match single-precision (FP32) accuracy and provide significant cost reduction, however, they are still over-provisioned and leave potential unexploited. There has been \textit{limited} success at using very small datatypes with 8b~\cite{IBM_8bit} and 4b~\cite{IBM_4bit} extremes for \textit{some} cases. Similarly, major hardware manufacturers are investigating how to use narrower floating point with different mantissa/exponent ratios according to perceived needs of tensors~\cite{fp8_Nvidia_Intel_Arm, rouhani2023microscaling}. These datatypes are often tailored to specific network architectures and current selection approaches cannot match FP32 accuracy outside of a small subset of shallow networks. Other energy efficient datatypes have been proposed including dynamic floating-point~\cite{DBLP:conf/iclr/0002MMKAB0VKGHD18}, flexpoint~\cite{Koster:2017:FAN:3294771.3294937}, hybrid block floating-point~\cite{Drumond:2018:TDH:3326943.3326985,MX_block_floating_point} and combinations with others like fixed-point~\cite{DBLP:conf/iclr/0002MMKAB0VKGHD18,mixedP,nvidia_mixedP}.
 
These \textit{tailored} methods require careful trial-and-error investigation of \textit{where}, \textit{when}, and \textit{which} datatypes to use. This is challenging because different tensors, tasks, architectures, or layers \textit{require} different datatypes.  The methods require full trial-and-error training runs and \textit{post mortem} analysis as whether the choice of datatypes is viable. Moreover, since the datatypes are statically chosen they offer no opportunity to amend the choice if accuracy suffers (e.g., significant drop with deeper networks~\cite{IBM_4bit}).

It's important to recognize that for preselected datatype methods, \textit{two} key decisions have to be made: 1) the meaning of the bits, and 2) the required number of bits. The first decision — defining the meaning of the bits — has usually been the bigger contribution. It involves choosing between options like integer or exponential representations, shared or individual exponents for floating points, or using lookup tables. Traditionally, the number of bits is determined through experience or experimentation. Our methods automate this selection process for any chosen representation. Moreover, the principles from \DPRL are independent of the bit meaning and can determine the optimal number and type of bits required. This enhances the process with automation, adaptability, and granularity, moving beyond the conventional, fixed approach.

Adaptable methods are gathering attention. \textit{Open-loop} methods modify the datatype based on a predetermined schedule but require trail-and-error runs to find an adequate schedule. \textit{Closed-loop} solutions that monitor some metric other than loss or task accuracy (e.g., quantization error) comparing against a preset allowable error schedule (based on time, layer depth, or other network features) run into the same issue~\cite{FASTDNN, CVPR_FixedPointBackPropagationTraining}.

ACGC~\cite{evans2021acgc} determines leaner datatypes to use in mixed-precision fixed-point quantization for \textit{activations}. It periodically determines the maximum permissible quantization error bound for each \textit{activation} tensor based on a user-selected maximum allowable increase in loss and adjusts the bitlength they use. ACGC can not compress weights and is not applicable where weights dominate such as most natural language processing networks. Determining the permissible bounds is also expensive, however, its overhead is kept down by performing it infrequently. 

Obviously, knowing in advance which compact datatypes to use, and when, would be the best. However, given that this goal still eludes us, our work asks whether we can harness the training process itself to automatically \textit{learn} them, 1)~\textit{automatically} tailoring datatypes to each tensor, layer, and network, and 2)~continuously adjusting them as training progresses, adapting to the changing needs.

We present a fully automatic closed-loop solution that tracks the loss. Our approach redefines mantissa and exponent quantization to make them differentiable and includes the reduction of datatype size as part of the objective of gradient descent, without high overhead. 

Effective closed-loop solutions for finding the most efficient datatype exist for \textit{inference}. They use reinforcement learning~\cite{HAQ}, differentiable datatype definitions~\cite{bitpruning,SDQ}, architecture search~\cite{DNAS_Quantization}, learnable parameters for every weight bit~\cite{yang2021bsq}, and profiling~\cite{DBLP:conf/iiswc/NikolicMM18}, etc., and have been proposed for fixed-point \textit{inference} mixed precision quantization. However, all of these are too expensive for training and their overheads would overshadow the benefits of a more compact training datatype. Moreover, some are specifically targeting weights or activations, and can not adapt to different architectures where the main footprint contributors may change (weight vs activation heavy cases). 

\section{Adjusting Value Containers}
\label{sec:bitpruning}

In general, maintaining accuracy on most real-world tasks requires floating-point-based training. These formats comprise a sign S, a mantissa M, and an exponent E:
 \begin{equation}
 V(S,M,E)= (-1)^{S} \times (1 + M) \times  2^{E}
  \end{equation}
Each part is differently distributed and requires unique approaches to effectively compress. 
The sign S only needs 1 bit and when V is limited to only positive numbers, it can be omitted. M, including its implied one, is the fractional part of the multiplier and, denormals aside, has a range $[1,2)$. Reducing M's length reduces the \textit{precision} of the full value. 
Finally, E is the exponent of the second multiplier. Reducing E's length narrows the \textit{range} of the full value:
 \begin{equation}
 V(S,M,E) \in [-V_{max}, -V_{min}] \cup \{0\}\cup [V_{min}, V_{max}] 
 \label{eq:exponent_range}
  \end{equation}
where $V_{max}$ and $V_{min}$ are the absolute values of the limits of V with the exponent range $[E_{min}, E_{max}]$ and maximum mantissa $M_{max}$:

\begin{equation}
V_{max} = (1 + M_{max}) \times 2^{E_{max}}
 \end{equation}

\begin{equation}
 V_{min} = 2^{E_{min}}
\end{equation}

Sections~\ref{sec:QMQE} and ~\ref{sec:mantissa_mc} present respectively machine learning and hardware-inspired approaches for learning mantissa and exponent bitlengths. Both methods omit the sign bit whenever possible. Section~\ref{sec:exponent} complements either approach with an optional lossless exponent compression method. 

We study \DPRL with ResNet18, ResNet50 ~\cite{resnet} and MobileNet V2~\cite{sandler2018mobilenetv2} trained on ImageNet~\cite{imagenet}, DLRM~\cite{DLRM19} trained on Kaggle Criteo, Vision Transformer~\cite{vit} pretrained on Cifar10~\cite{cifar10}, BERT~\cite{BERT} finetuned on GLUE~\cite{Glue} and GPT--2~\cite{GPT2} finetuned on Wikitext 2~\cite{wikitext2}.
We report detailed results with ResNet18 and conclude with overall results for all models.

\subsection{Machine Learning Approach}
\label{sec:QMQE}

\MP and \QE learn mantissa and exponent bitlengths, respectively. Both use inexpensive procedures for both the forward and the backward pass of training and rely on making quantization differentiable and penalizing the larger bitlengths in the loss function. We begin by defining a conventional quantization scheme for integer mantissa and exponent bitlengths in the forward pass, and then expand it to the non-integer domain to allow gradient descent to learn bitlengths. A parameterizable loss function guides this learning by penalizing larger bitlengths. We then touch on the compute overhead of our methods and the plan for the final selection of mantissa bitlengths. Ultimately, we demonstrate the benefits of this approach on memory footprint during ImageNet training. 

\noindent\textbf{\MP (\QMshort): }
The greatest challenge for learning bitlengths is that they represent discrete values with no obvious differentiation. To overcome this, we define our quantization for non-integer bitlengths, starting with an \textit{integer quantization} of the mantissa $M$ with $n_m$ bits by removing all but the top $n_m$ bits:
 \begin{equation}
 P(M,n_m)= M \wedge (2^{n_m}-1)<<(m-n_m)
 \label{eq:mantissa}
  \end{equation}
where $P(M,n_m)$ is the mantissa with bitlength $n_m$, $m$ the maximum bitlength and $\wedge$ a bitwise AND.

This scheme does not allow the learning of bitlengths with gradient descent due to its discontinuous and non-differentiable nature. To expand the definition to real-valued $n_m = \floor{n_m}+ \{n_m\}$, the values used in inference during training are \textit{stochastically} selected between the nearest two integers with probabilities $\{n_m\}$ and $1 - \{n_m\}$:
\begingroup
\thickmuskip=1mu
\medmuskip=0mu 
\begin{equation}
     P(M,n_m)=
    \begin{cases}
      P(M,\floor{n_m}),&\text{w/ prob.}\ 1 - \{n_m\} \\
      P(M,\floor{n_m}+1),&\text{w/ prob.}\ \{n_m\} 
    \end{cases} \label{eq:stochastic}
  \end{equation}
\endgroup
where $\floor{n_m}$ and $\{n_m\}$ are floor and fractional parts of $n_m$.

This mantissa approach faithfully represents the relationship between bitlength and precision in an \textit{inexpensive} way. The overhead is limited to the single bitlength parameter and a random number (in the forward pass) per value group (e.g., a tensor), and a single multiply-accumulate operation (in the backward pass) per value.

\noindent\textbf{\QE (\QEshort): }
The exponent range is parameterized as follows: 

\begingroup
\thickmuskip=1mu
\medmuskip=0mu
\thinmuskip=0mu 
\begin{equation}
 R(V, V_{max}, V_{min})=
 \begin{cases}
      -V_{max}, & V \in (-\infty,-V_{max})   \\
      V, & V \in [-V_{max},-V_{min}]\\ 
      -V_{min}, & V \in (-V_{min},-V_{min} / 2]\\ 
    0, & V \in (-V_{min} /2, V_{min} / 2) \\
      V_{min}, &  V \in [V_{min} / 2,V_{min})\\
      V, & V \in [V_{min}, V_{max}]\\
      V_{max}, & V \in (V_{max}, \infty)  \\
    \end{cases}
  \end{equation}
\endgroup
where $V_{max}$ and $V_{min}$ are boundaries from Equation~\ref{eq:exponent_range}.

The partial derivatives of this function with respect to $V$, $V_{max}$ and $V_{min}$ are:
\begin{equation}
\frac{\partial R}{\partial V} = 
\begin{cases}
      0, & V \in (-\infty,-V_{max}]  \\
      1, &  V \in (-V_{max}, V_{max})\\ 
    0, & V \in [V_{max}, \infty) 
    \end{cases}
  \end{equation}
\begin{equation}
\frac{\partial R}{\partial V_{max}} = 
\begin{cases}
      -1, &V \in (-\infty,-V_{max}]  \\
      0, & V \in (-V_{max}, V_{max})\\ 
    1, &  V \in [V_{max}, \infty)   
    \end{cases}
  \end{equation}
\begin{equation}
\frac{\partial R}{\partial V_{min}} = 
\begin{cases}
      0, & V \in (-\infty,-V_{min}]  \\
      -1, & V \in (-V_{min},-V_{min} / 2]\\
      1, & V \in (-V_{min} /2, 0) \\
      -1, & V \in [(0, V_{min} / 2) \\
      1, &  V \in [V_{min} / 2,V_{min})\\
    0, & V_{min} < V    
    \end{cases}
  \end{equation}
The next challenge of finding the exponent bitlength gradient is to connect the value range with the exponent range:
\noindent
\begingroup
\thickmuskip=1mu
\medmuskip=0mu
\thinmuskip=0mu \begin{equation}
V_{max} = (1 + M_{max}) \times 2^{E_{max}}
\label{eq:V_max}
 \end{equation}
\endgroup
\begingroup
\thickmuskip=1mu
\medmuskip=0mu
\thinmuskip=0mu 
\begin{equation}
 V_{min} = 2^{E_{min}}
\label{eq:V_min} 
\end{equation}
\endgroup

Where $M_{max}$ is the largest possible mantissa, $E_{max}$ is the largest possible exponent, and $E_{min}$ is the smallest possible exponent. For simplicity, we choose our exponent range to be symmetrical around 0:  
\noindent\
\begin{equation}
E_{min} = - 2^{n_e^i-1}
\end{equation}
\begin{equation}
E_{max} = 2^{n_e^i-1} -1
\end{equation}

where the integer $n_e^i$ is the integer exponent bitlength. The bias can also be learned, however, this is not essential as the important exponents will be around 0. As with \QMshort, we expand this definition to the continuous domain stochastically:
\begin{equation}
n_e^i = 
\begin{cases}
      \floor{n_e}, & \text{w/ prob.}\ 1 - \{n_e\} \\
      \floor{n_e}+1, & \text{w/ prob.}\ \{n_e\} 
    \end{cases}\label{eq:expstochastic}
 \end{equation}
where ${n_e}$ is the learnable exponent bitlength.

Similar to \QMshort, this approach faithfully represents the relationship between exponent bitlength and the range in an \textit{inexpensive} way. Its overhead is limited to the single bitlength parameter and a random number (in the forward pass) per value group sharing a datatype (e.g., tensor), and a single operation (in the backward pass) per value.

Finally, in order to obtain the fully quantized value we first bound the input with $R$ to remove the exponent bits and then apply $P$ to remove the mantissa bits.

\noindent\textbf{Datatype Learning: }
These schemes are applied to each activation and weight tensor separately. Since the minimum bitlength is 0, $n_m$ and $n_e$ are clipped at 0. This extension of the bitlengths in the continuous domain allows the loss to be differentiable with respect to both E and M bitlengths.

The formulae above are applied during the forward pass. Quantized values are saved and used in the backward pass. This strategy reduces the footprint of training because only quantized values are used in forward and backward passes.  

\noindent\textbf{Loss Function: }
We augment the loss $L$ to penalize M and E bitlengths by adding a weighted average of their volume:
 \begin{equation}
    L=L_l+{\gamma}_m \times \sum^{i}(\lambda_i \times n_m^i) +{\gamma}_e \times \sum^{i}(\lambda_i \times n_e^i)
  \end{equation}
where $L_l$ is the original loss, ${\gamma}_m$ and ${\gamma}_e$ are regularization coefficients determining  quantization aggressiveness, $\lambda_i$ is the importance of the $i^{th}$ group of values (one per tensor), and $n_m^i$ and $n_e^i$ are the mantissa and exponent bitlengths of the activations or weights in that tensor. 

\noindent\textbf{Competing Objectives: }Our augmented loss adds a competing objective for training. Overemphasizing bitlength choice may sacrifice task performance, while underemphasizing it may sacrifice potential gains. Balancing the objectives via $\gamma$ selection proves straightforward for two reasons. First, from our experience, selecting $\gamma$ such that the bitlength loss component is 1-2 orders of magnitude smaller than the main task objective loss is enough to squeeze out most of the reduced datatype benefits whilst being sufficiently small to \textit{not} adversely influence final accuracy. Second, finding the best $\gamma$ isn't necessary since learning the bitlengths is a very coarse task, and at the end, the bitlengths have to be rounded to appropriate integer ones. For all experiments setting both $\gamma_m$ and $\gamma_e$ to $0.1$ proved sufficient.

\noindent\textbf{Target Criteria: }Our loss function can target any quantifiable criteria by a suitable selection of $\lambda_i$'s. Since our goal is to minimize the total footprint of training, we weigh each tensor according to its memory footprint. 

\noindent\textbf{Overhead: } 
\QMshort and \QEshort add minimal computational and memory overheads.
In the forward pass, random numbers are needed at a chosen granularity as per eq.~\ref{eq:stochastic} and~\ref{eq:expstochastic}. Our experiments show that a random number per tensor per batch is sufficient and is of a negligible cost. 

To update the bitlength parameters in the backward pass, we need to compute their gradients. These are a function of the quantized values and gradients, which are calculated during the regular backward pass. The extra calculations are proportional to the number of values. This overhead is negligible in comparison to the total number of computations. For instance, for ResNet18 the overhead is less than $1\%$.

The only new parameters that are stashed are the four floats per layer (mantissa and exponent bitlength for weights and activations), negligible in comparison with the total footprint. All other values are consumed as they are produced.

\noindent\textbf{Bitlength Selection Schedule: }
\QMQEshort use non-integer bitlengths. We arrive at integer bitlengths, by disabling learning bitlengths when they are not needed at which point we \textit{round up} the bitlengths and freeze them. Our experiments show that bitlengths converge quickly to the final ones within a couple of epochs. Accordingly, we freeze the bitlengths after epoch 5. This avoids the small overhead for most of the training. In our experiments, we found that when both methods are used concurrently some tensors \textit{might} need more bits later in training. Accordingly, we re-enable \QMshort and \QEshort for an additional 5 epochs on every learning rate change. This allows precision to increase where necessary to accommodate the reduction in update magnitudes. Regardless of whether \QMshort and \QEshort are enabled or disabled, the benefits of reduced bitlengths apply throughout training. This "fancy" schedule is not fully needed. Experiments where we fixed the bitlengths after 5 epochs still converged and achieved \textit{slightly} lower accuracy.

\begin{figure}[t!]
\centering
\subfloat[Validation accuracy]{
\includegraphics[width=0.9\columnwidth]{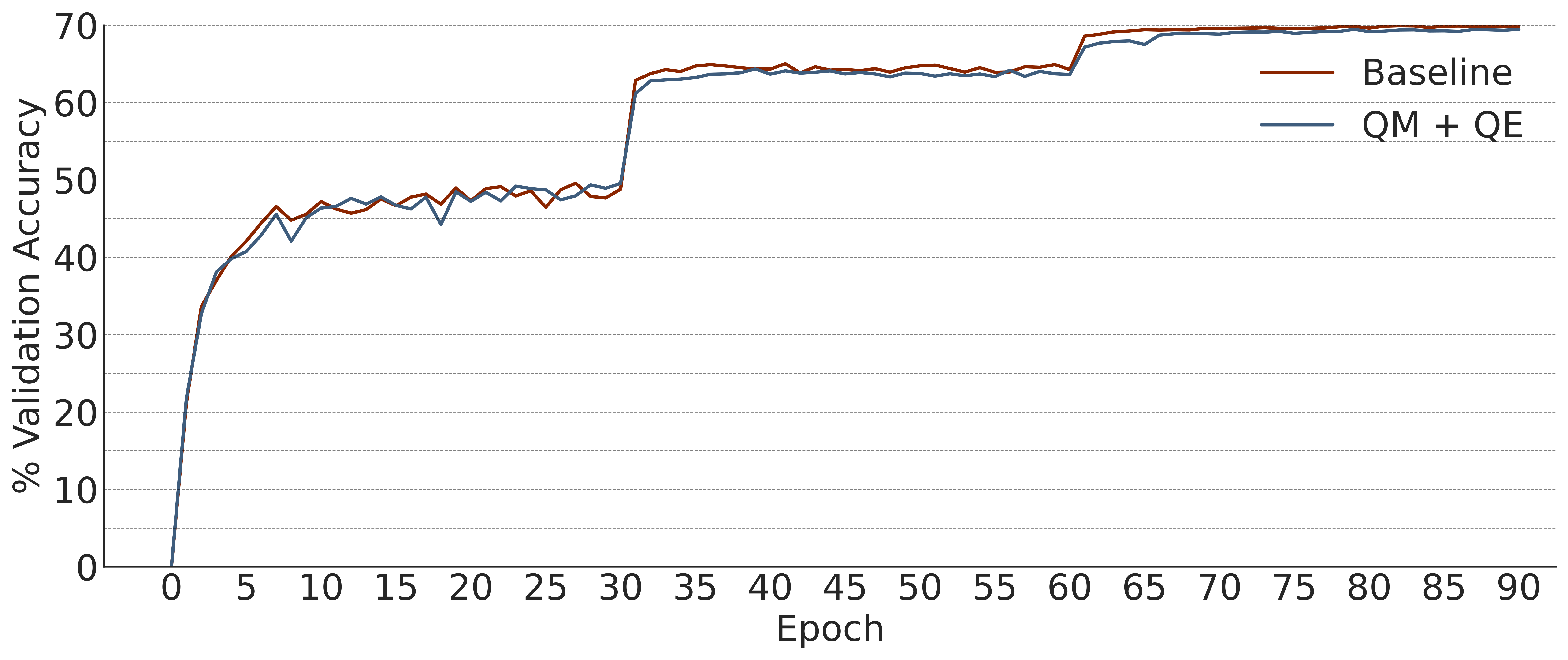}
\label{fig:resnet18_accuracy}
}

\subfloat[Mantissa bitlength]{
\includegraphics[width=0.9\columnwidth]{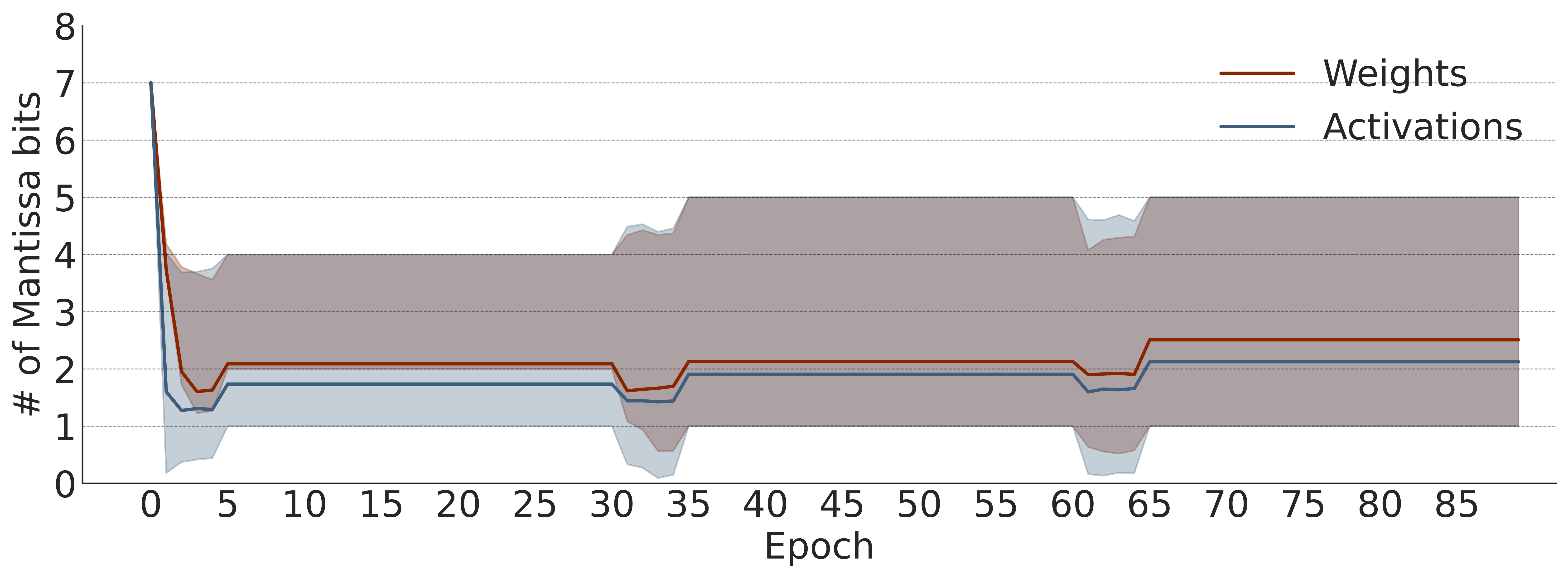}
\label{fig:resnet18_bitlengths_man}
}

\subfloat[Exponent bitlength]{
\includegraphics[width=0.9\columnwidth]{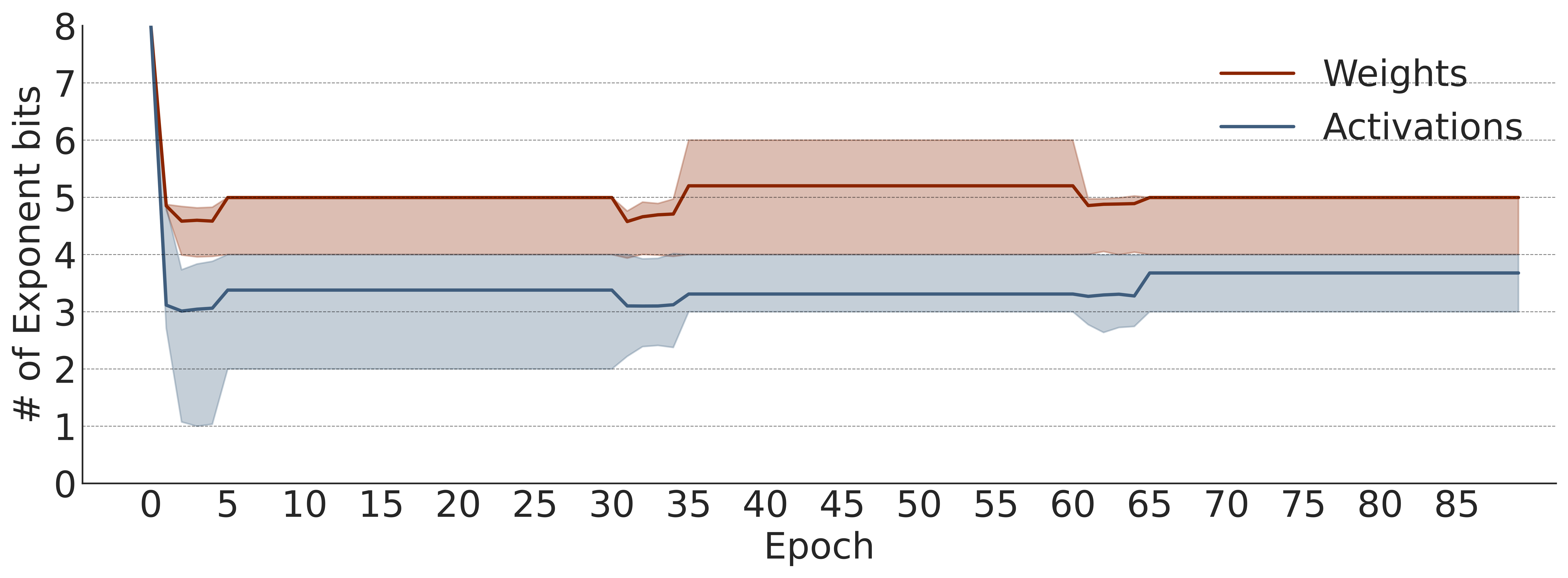}
\label{fig:resnet18_bitlengths_exp}
}

\subfloat[Total bitlength]{
\includegraphics[width=0.9\columnwidth]{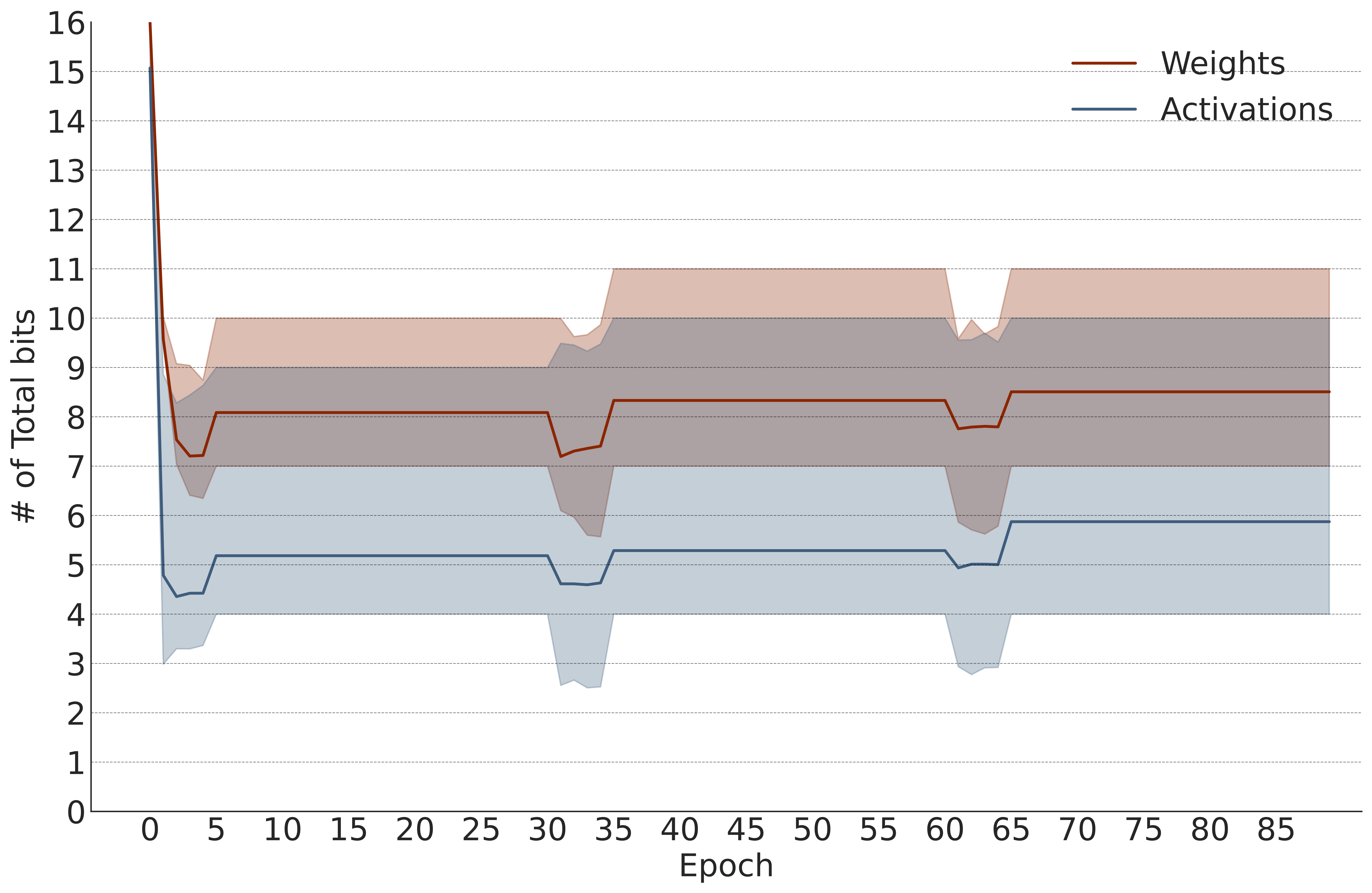}
\label{fig:resnet18_bitlengths_total}
}

\caption{\QMshort and \QEshort on ResNet18/ImageNet throughout training: (a) Validation accuracy, (b) Weighted mantissa bitlengths with their spread, (c) Weighted exponent bitlengths with their spread, and (d) Weighted total bitlengths with their spread.}
\end{figure}

\noindent\textbf{Evaluation: Bitlengths and Accuracy: }
\label{sec:experiments}
We report measurements for per-layer weights and activations quantized separately using a loss function weighted to minimize overall memory footprint. We train ResNet18 on the ImageNet dataset over 90 epochs, with regularizer strength of 0.1, learning rate of 0.1, 0.01 and 0.001 respectively at epochs 0, 30, and 60 and weight decay of 0.0001. Both \QMshort and \QEshort excel at minimizing the memory footprint whilst not introducing accuracy loss. Figure~\ref{fig:resnet18_accuracy} shows that throughout training, our methods introduce minimal changes in validation accuracy converging to a solution within $0.4\%$ of the FP32 baseline. Minor accuracy loss occurs when the methods are actively pushing bitlengths to their limits. Any loss is quickly regained when bitlenghts are frozen and rounded up since this relaxes the value range.

\noindent\textbf{\QMshort: }Figure~\ref{fig:resnet18_bitlengths_man} shows how \QMshort quickly (within a couple of epochs) reduces mantissas below $2$b on average. The large spread in bitlengths across layers shows that \QMshort's granular, per tensor approach is the right choice for boosting benefits. In comparison, FP8 would use $2\text{-}3$b (out of 8) everywhere~\cite{fp8_Nvidia_Intel_Arm}. While \QMshort \textit{sometimes} allocates more than 3b for some tensors, this slack boosts overall footprint reduction since it enables shorter bitlengths for larger tensors. Finally, the results show a minor increase of bitlengths across period boundaries of our bitlength learning schedule. The total training cumulative mantissa footprint is reduced to $8.4\%$ of the FP32 mantissa footprint ($8.3\%$ for activations and $9.8\%$ for weights). 

\noindent\textbf{\QEshort: }Figure~\ref{fig:resnet18_bitlengths_exp} shows that learning exponent bitlengths via \QEshort exhibits similar behavior.  Bitlengths quickly converge to $4$b or less for activations, and on average down to around $5$b for weights. In comparison, FP8 would use $4\text{-}5$b (out of 8) everywhere, a fair choice for network-wide bitlength~\cite{fp8_Nvidia_Intel_Arm}. \QEshort sometimes uses longer exponents for some tensors enabling short exponents for large tensors. As a result, \QEshort outperforms FP8 in exponent footprint. Compared to mantissas, the spread in exponent bitlengths across layers is lower yet significant while there is a more noticeable increase of bitlengths from one learning period to the next. The cumulative training memory footprint is reduced to $43.1\%$ of the FP32 exponent footprint ($42.7\%$ for activations and $62.8\%$ for weights). 

\noindent\textbf{\QMQEshort: }Figure~\ref{fig:resnet18_bitlengths_total} shows the total bitlength of the datatype for each tensor, including sign, mantissa, and exponent. It further amplifies the conclusions from above. Massive footprint reduction, significantly varying bitlength tensor to tensor justifying the fine-grained approach, and slightly increasing bitlength for some tensors learning period to period. To further emphasize the importance of the fine-grained approach we can look at the average and worst-case bitlengths. For instance, the worst-case activation tensor requires 11b while the average is less than 6b.

\begin{figure}[t!]
\centering
\includegraphics[width=0.9\columnwidth]{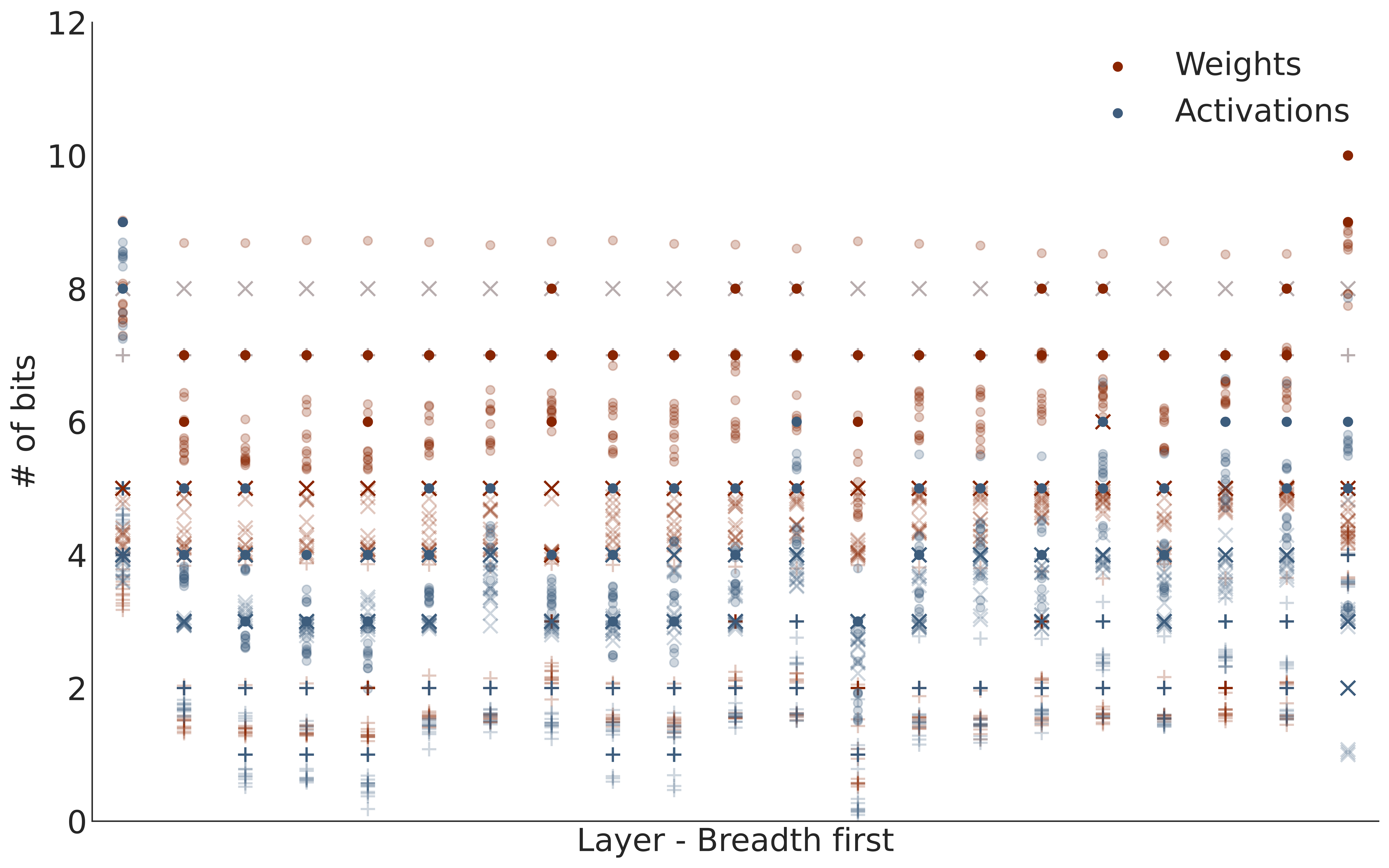}
\caption{\MP and \QE on ResNet18/ImageNet: mantissa ($+$), exponent ($\times$), and total ($\cdot$) bitlength datatypes of each tensor at the end of each epoch. Darker colors indicate multiple occurrences.}
\label{fig:resnet18_bitlenghts_per_layer}

\end{figure}

The variability of total, exponent and mantissa bitlenghts for weights and activations at the beginning of every epoch is shown in Figure~\ref{fig:resnet18_bitlenghts_per_layer}. This figure shows that, while there are some tensors that share bitlenghts, for instance, weight exponents, most bitlengths vary wildly. The most important message from this graph though is that choosing datatypes is hard and complicated. If we want to squeeze as much of a reduction of footprint as possible, we need an automated method. It is impossible to guess the bitlengths in advance.

\begin{figure}[]
\centering
\includegraphics[width=.30\linewidth]{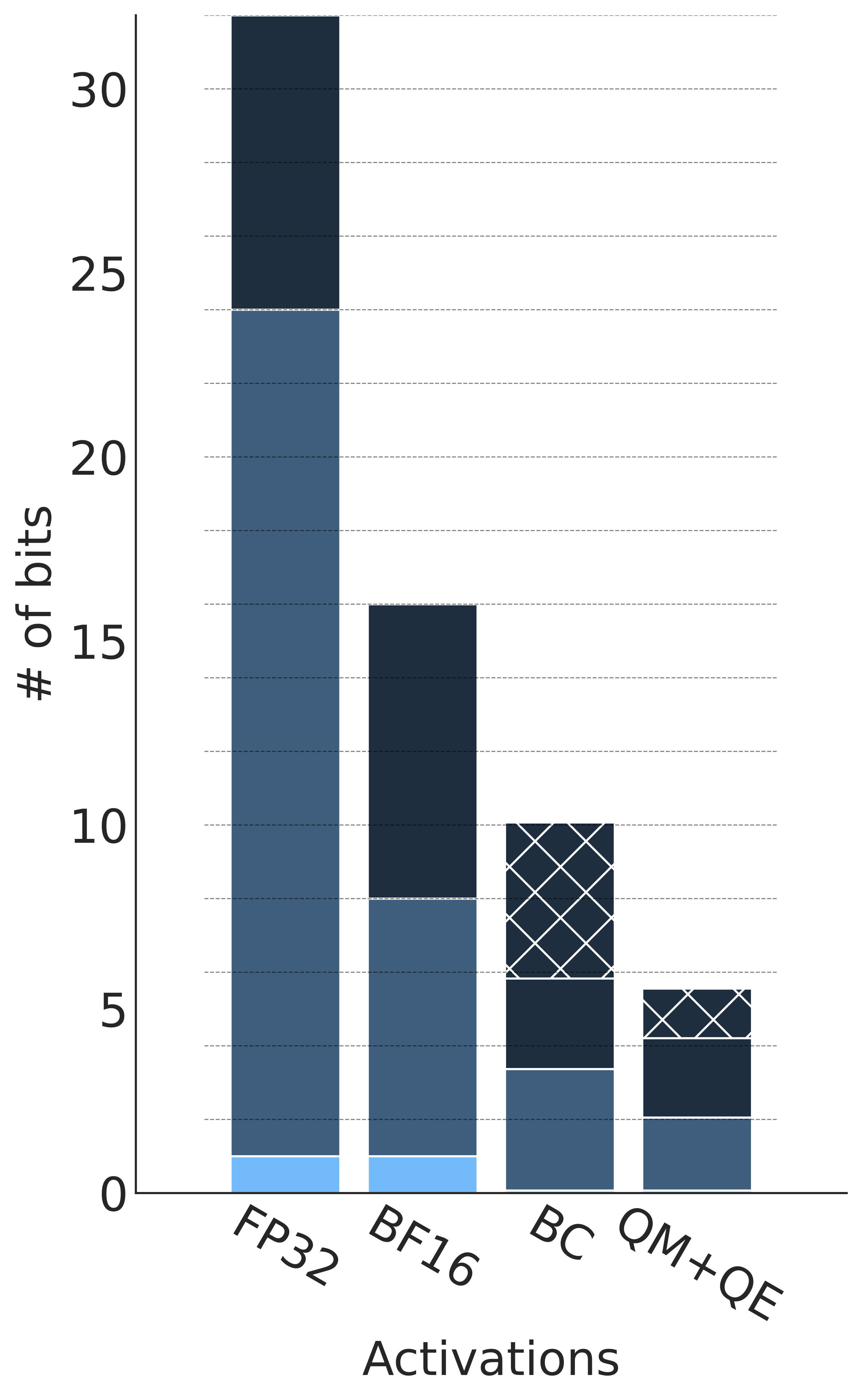}
\includegraphics[width=.30\linewidth]{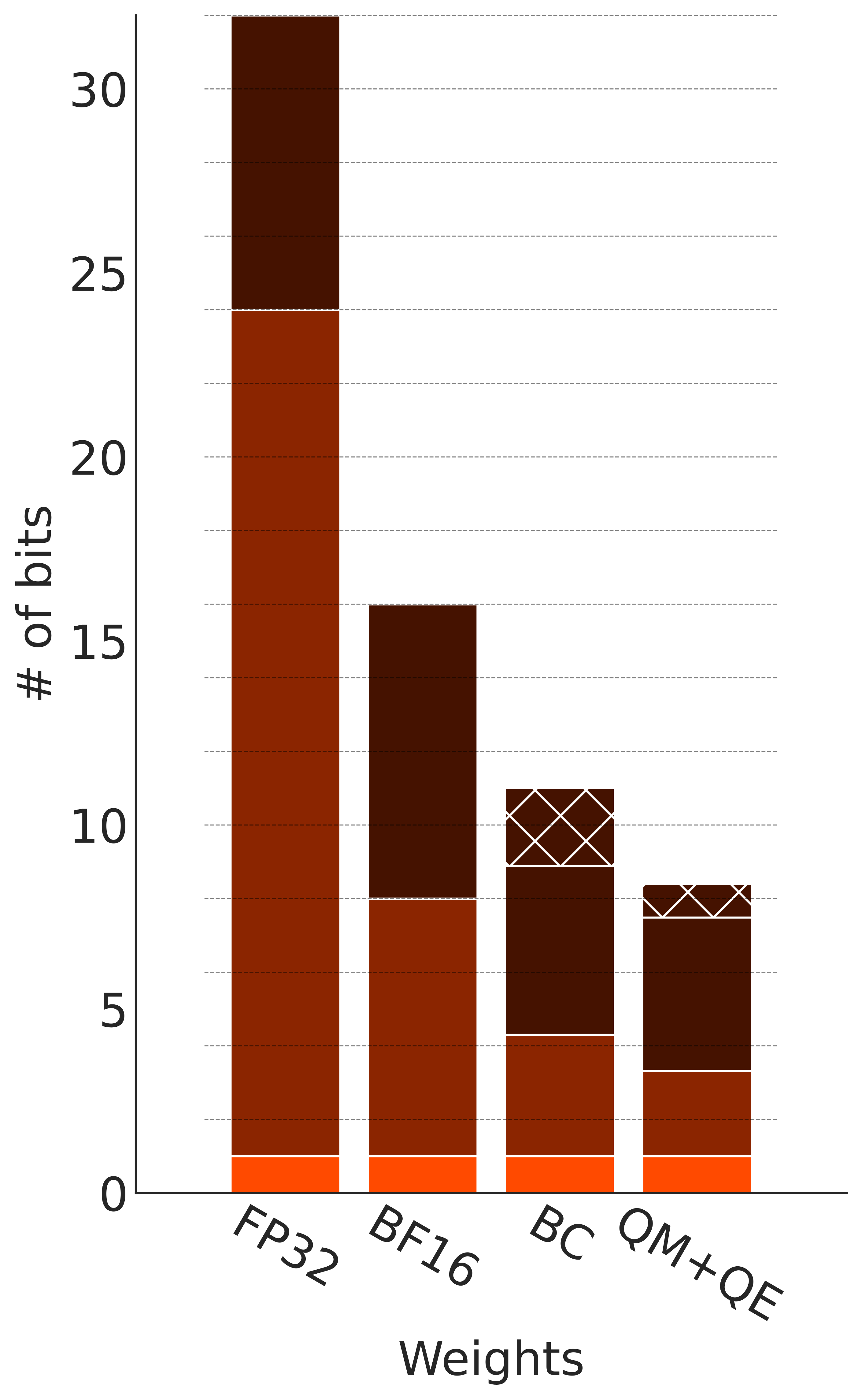}
\includegraphics[width=.30\linewidth]{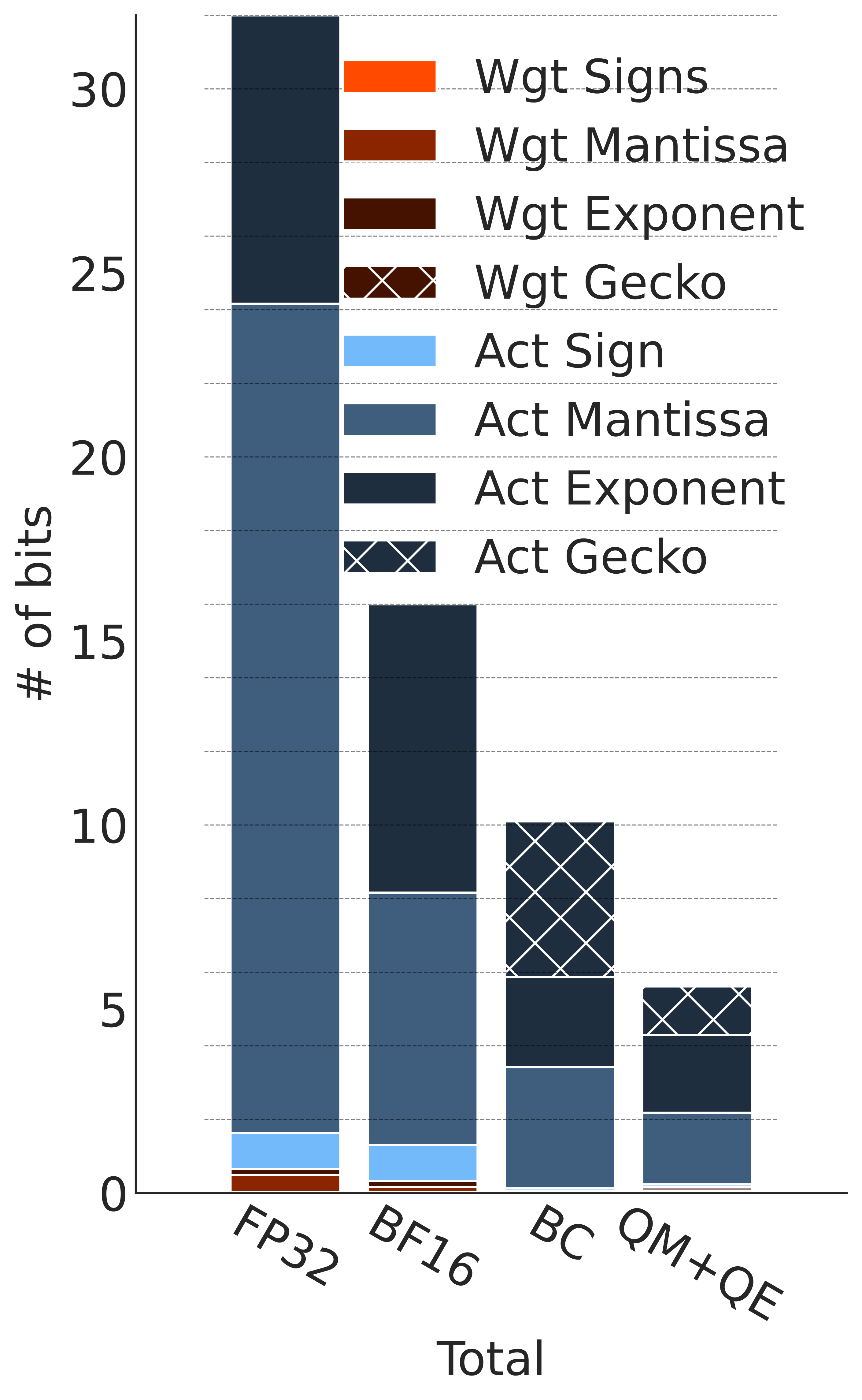}

\caption{\DPRL: Relative training footprint of ResNet18 with FP32, BFloat16, \GMC and \GMP.}
\label{fig:resnet18_mem_reduction}
\end{figure}

Cumulatively, on average, the datatype footprint is reduced by $3.86\times$ ($8.28$ bits), $5.92\times$ ($5.40$ bits) and $5.86\times$ ($5.46$ bits) vs FP32 for weights, activations, and total footprint. Similarly, footprint reduction in comparison with BFloat16 is $1.93\times$, $2.96\times$, and $2.93\times$ for weights, activations, and total, respectively. This is further emphasized in Figure~\ref{fig:resnet18_mem_reduction}. Finally, \QMQEshort datatype is $32\%$ smaller than FP8.

\noindent\textbf{\QMQEshort as a datatype selection advisor:} \QMQEshort quickly learns bitlengths that can be used to learn the per tensor datatypes to use for training the network, e.g., if we need to retrain the network we can use bitlenghts from the previous run as-is. The accuracy of such a run increased $0.2\%$ of previous training with \QMQEshort. Similarly, bitlenghts learned in the first 5 epochs can be used with a small accuracy drop~($0.7\%$). This capability is particularly useful given that industry is introducing a wide selection of leaner datatypes. It can aid or completely replace the current, manual, trial-and-error selection process allowing users to automatically benefit from the datatypes their hardware supports. 

\subsection{\MC}
\label{sec:mantissa_mc}
\begin{figure}[t]
\centering
\subfloat[Validation accuracy]{
\includegraphics[width=0.85\columnwidth]{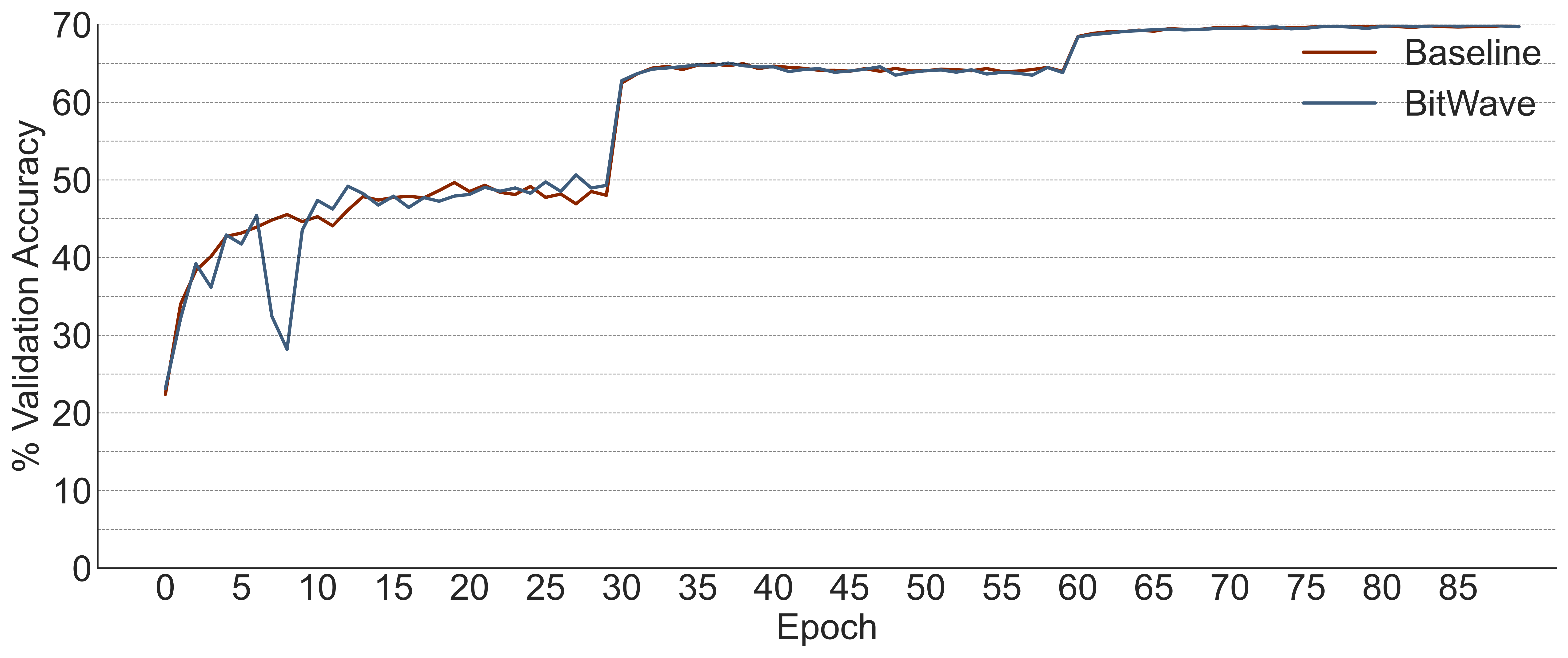}
\label{fig:resnet18_accuracy_bitchop}
}

\subfloat[Average bitlengths over time]{
\includegraphics[width=0.85\columnwidth]{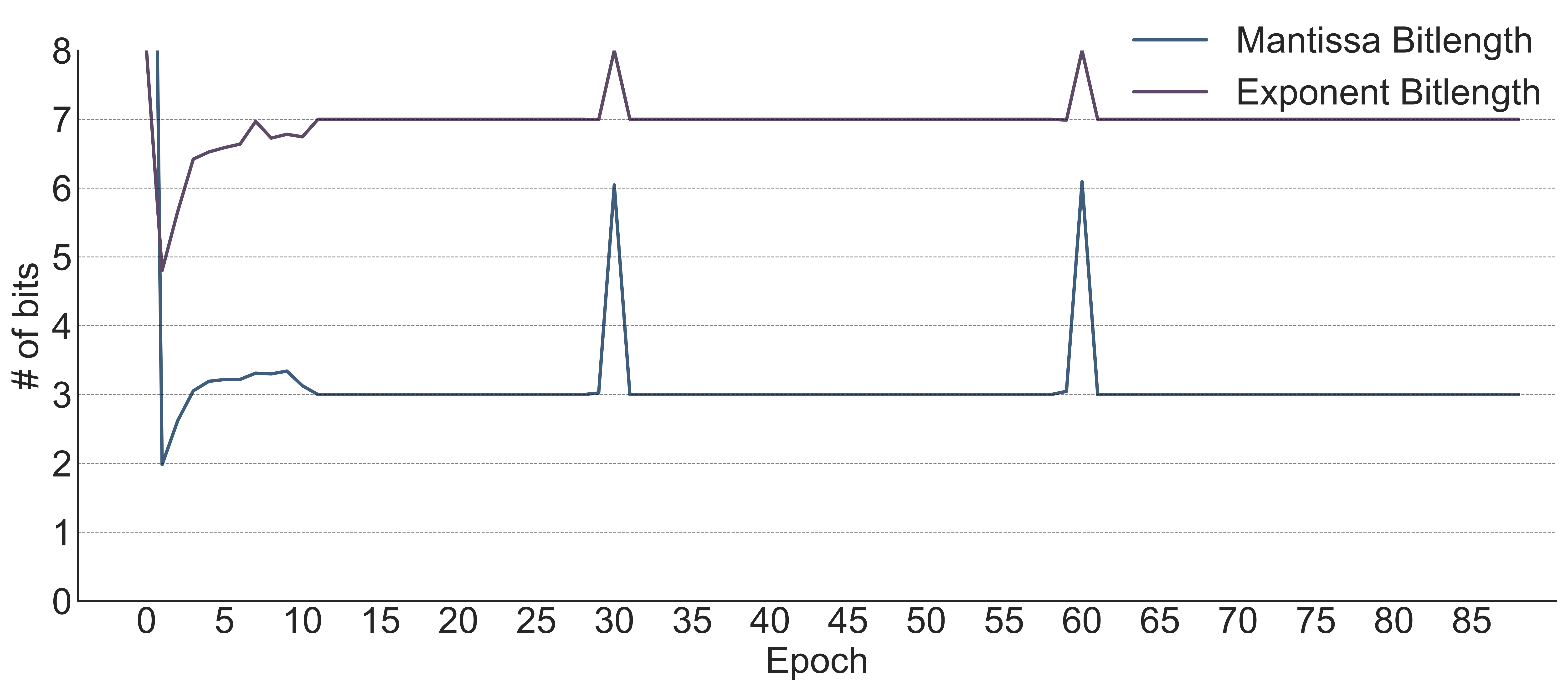}
\label{fig:resnet18_bitlenghts_per_epoch_bitchop}
}

\subfloat[Distribution of mantissa bitlengths]{
\includegraphics[width=0.85\columnwidth]{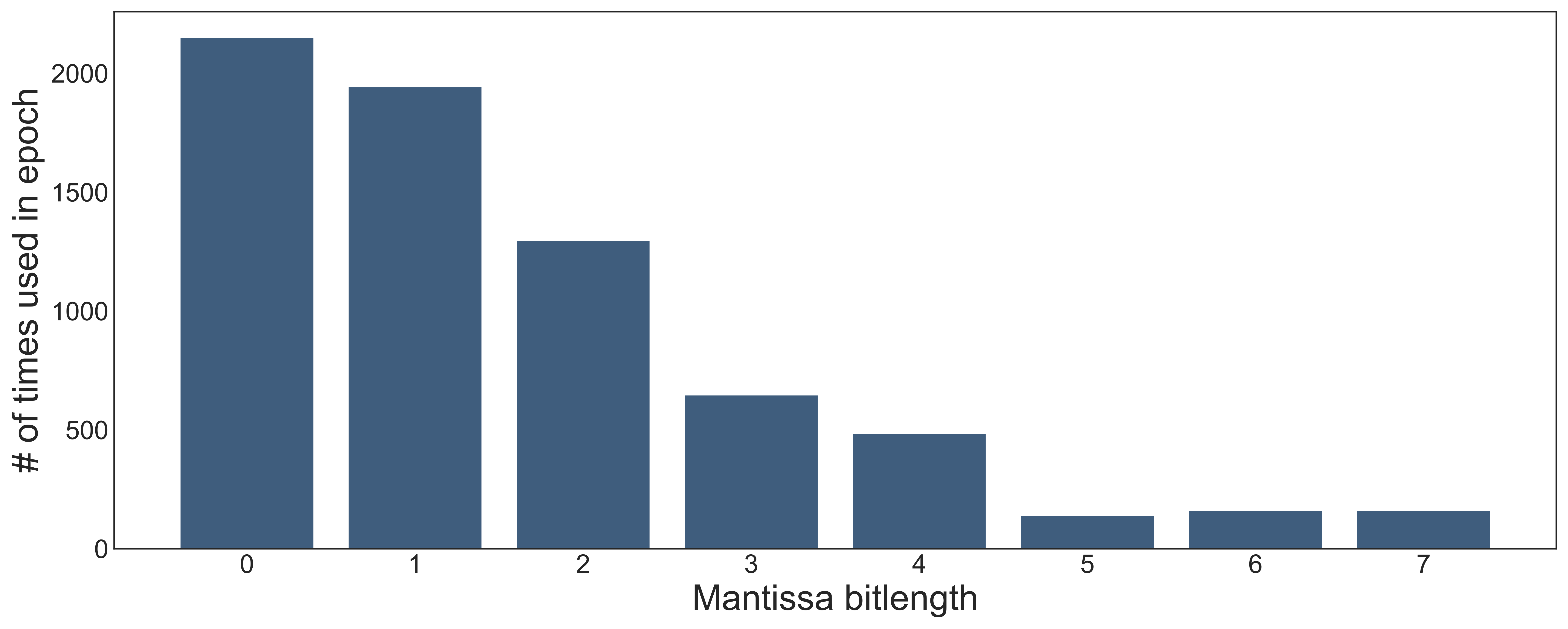}
\label{fig:resnet18_bitlenghts_histogram_bitchop}
}

\caption{\MC on ResNet18/ImageNet: (a) Validation accuracy throughout training, (b) Average mantissa and exponent bitlengths per epoch throughout training, (c) Distribution of \MC's mantissa bitlengths throughout the 5005 batches of epoch 5 of training.}
\end{figure}
Methods that do not interject into the training implementation, no matter how little, and that do not have any overhead are appealing. \MC is such a method. \MC monitors training progress as an outside observer adjusting the mantissa bitlength and exponent ranges accordingly: as long as the network is improving, \MC will attempt to use a shorter mantissa (\BWMshort) and to reduce the available exponent value range (\BWEshort). The ideal scenario for \MC is one where past observations of training progress are good indicators of forthcoming behavior. Fortunately, training is a long process based on \textit{trial-and-error}, which may be forgiving for momentary lapses in judgement. 

The main design decision that impacts how successful \MC will be is the information to use as a proxy for training progress. \MC should strike a balance between reducing bitlengths while avoiding over-clipping and hurting learning progress. We have experimented with several options and arrived at the following choices: 1)~Using the slope of a simple linear regression over a history of the loss as a proxy for network progress, 2)~observing training progress and adjusting bitlengths every batch, and 3)~using the same bitlengths for the entire network.

While the loss improves over time, at batch granularity it exhibits non-monotonic (sometimes erratic) behavior. \MC compensates for this by calculating a least squares regression (minimization of total sum of squared differences between predicted and history values) over a history of previous loss values. It uses the slope of the linear regression at each batch to smooth the non-monotonic behavior. 

\noindent\textbf{\MC \textit{Mantissa (\BWMshort):} }
\MC adjusts the mantissa length (unchanged, lower, or higher) by observing the slope of the linear regression. A negative slope indicates learning is improving, allowing for further mantissa trimming. A positive slope indicates no learning progress and \MC responds by increasing mantissa bitlength. If the slope is within a small threshold $T$ of 0.0, then \MC keeps observing and does not alter the bitlength.

\noindent\textbf{\MC \textit{Exponent (\BWEshort):} }
Considering the range of FP32 exponents ($[-126, 127]$), \MC adapts the range of values symmetrically by adjusting both limits. Exponents below the minimum are clamped to 0, whereas those above the maximum value saturate at that. This gradual change eventually reduces or increases the exponent bitlength. \MC adjusts the exponent range (unchanged, lower, or higher) by examining the slope of the calculated linear regression. A  negative slope  (with a threshold $T$) is assumed to indicate improvement, allowing the range to shrink. A positive slope (with the same threshold $T$), indicates deteriorating learning so \MC widens the exponent range. 

\noindent\textbf{Bitlength Selection Schedule: }
Similarly to \QMQEshort, \MC produces non-deterministic datatypes due to its intrinsic fluctuations throughout training because of its heuristic nature. To avoid this non-determinism and provide usable bitlengths for inference, \MC fixes the mantissa bitlength and the exponent range after a few epochs of training, by calculating the average of all the bitlengths up to that point of training, as well as the average of the exponent range. \MC then uses these averages for the rest of training. Experiments on the convergence of \MC show that the networks converge to the same accuracies ($\pm0.1\%$) whether the bitlengths are fixed or not, and there are evident benefits of creating deterministic inference-capable bitlengths.

\noindent\textbf{Evaluation: Bitlengths and Accuracy: }
We report \MC's effect on footprint and accuracy during training of ResNet18. Figure~\ref{fig:resnet18_accuracy_bitchop} shows that validation accuracy is unaffected. Figure~\ref{fig:resnet18_bitlenghts_per_epoch_bitchop} shows that \MC reduces mantissa bitlengths to 3b \textit{on average} from baseline precision. However, mantissa bitlengths may vary slightly per batch as illustrated in the histogram (Figure~\ref{fig:resnet18_bitlenghts_histogram_bitchop}) of bitlengths used throughout a sample epoch. This shows that training sometimes requires the entire range whereas other times it only requires 0 bits. Across the training process, \MC reduces the total mantissa footprint to $14.3\%$ of baseline, and the total exponent footprint to $83.8\%$. While \MC might miss bitlength reductions per layer and not reduce the exponent bitlength as much, it is non-intrusive and has no overhead.

\begin{figure}[t!]
\centering
\subfloat[]{
    \includegraphics[width=.85\columnwidth]{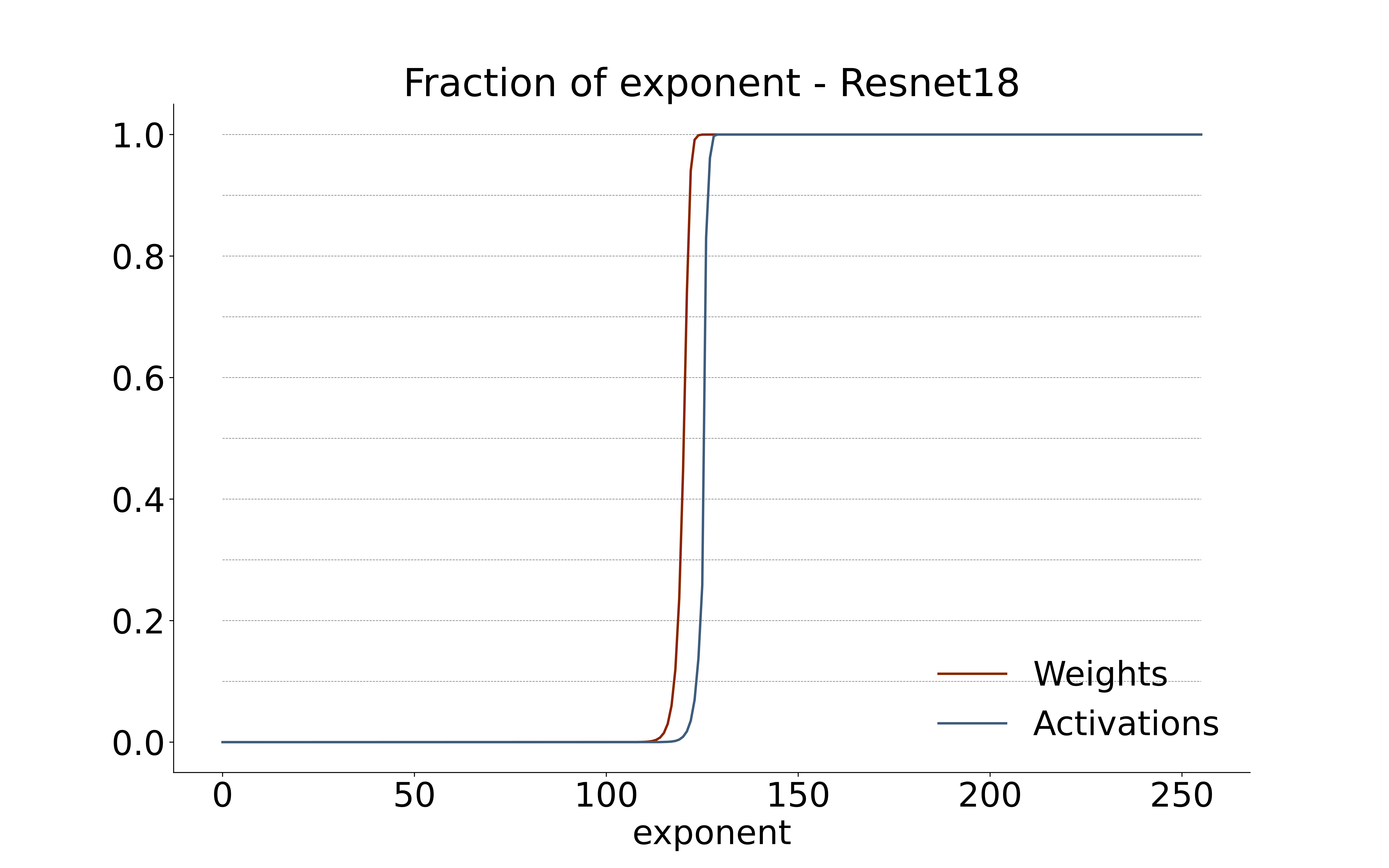}
    {\label{fig:original}}
    \label{fig:original_distribution}
}

\subfloat[]{
    \includegraphics[width=.85\columnwidth]{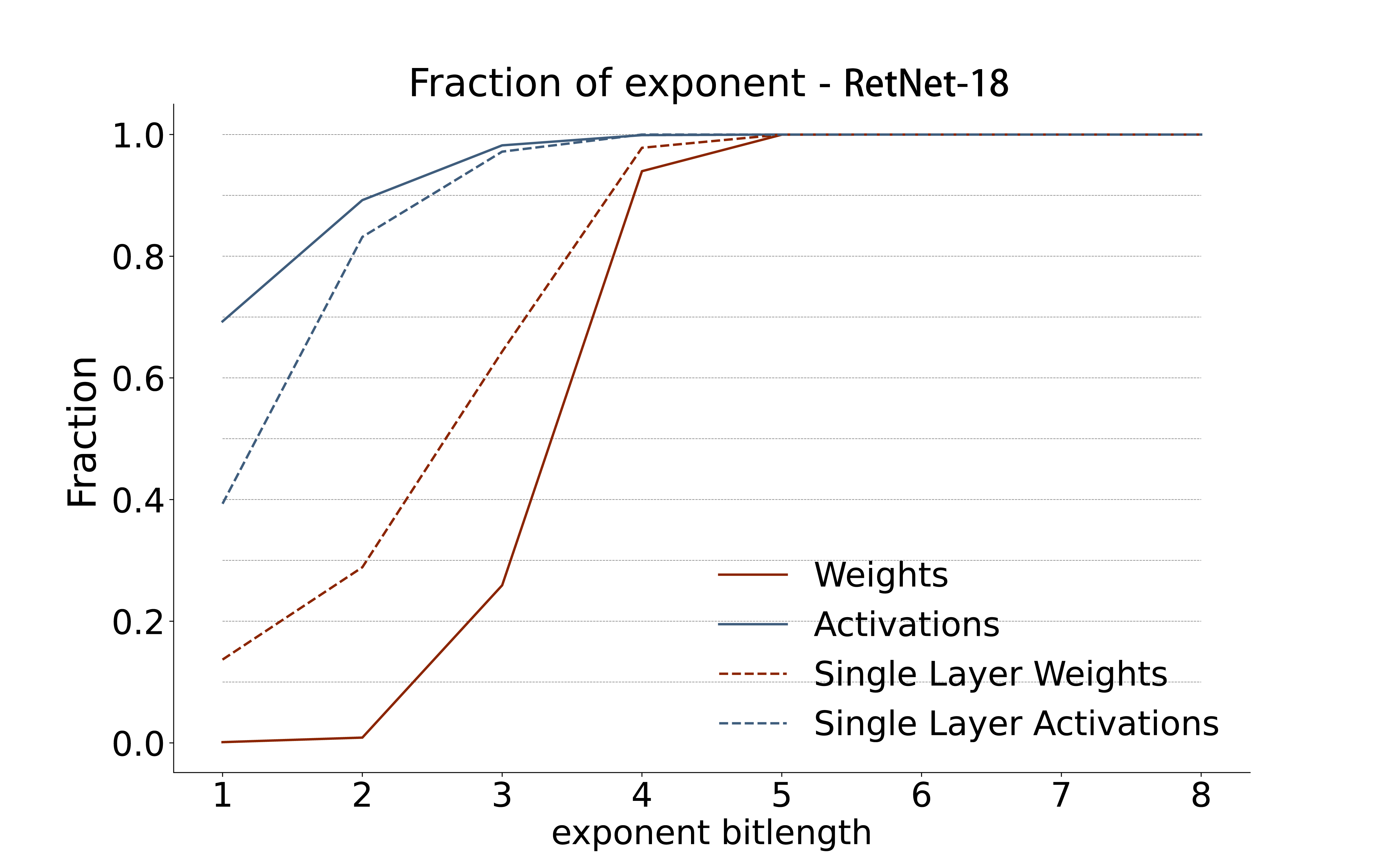}
    {\label{fig:compressed}}
    \label{fig:compressed_distribution}
}
\caption{\EC on ResNet18/ImageNet: (a) Cumulative distribution of exponent values. (b) Post-encoding cumulative distribution of exponent bitlength.}
\end{figure}

\begin{table*}
\centering
\caption{\GMC and \GMP: Validation metrics and total memory footprint reduction vs. FP32.}
\label{tbl:total_mem_reduction}
\resizebox{1\linewidth}{!}
{
\begin{tabular}{l|ll|c|c|ccc|cc|c|ccc|cc}
 & & & & \multicolumn{6}{c|}{\textbf{\GMP}} & \multicolumn{6}{c}{\textbf{\GMC}} \\
\textbf{Network}            &\textbf{Task} &\textbf{Metric} & \makecell{\textbf{FP32}\\\textbf{Score}} &\textbf{Score}  & \QMshort & \QEshort & \QEGshort      & \QMQEshort       & \makecell{\QMQEshort\\\Gshort}  &\textbf{Score}  & \BWMshort & \BWEshort & \BWEGshort      & \BWMBWEshort       & \makecell{\BWMBWEshort\\\Gshort}\\\hline
\textbf{ResNet18}       &Image Classification & Accuracy & 69.94 &69.50&$11.91\times$ & $2.322\times$& $3.649\times$ & $5.857\times$ & $7.599\times$ & 69.95& $6.961\times$& $1.193\times$& $3.235\times$& $3.197\times$& $5.539\times$\\
\textbf{ResNet50}       &Image Classification  & Accuracy & $76.06$ & 75.58&  $14.10\times$ & $2.277\times$ & $3.513\times$ & $6.192\times$ & $8.138\times$& 75.80& $7.385\times$& $1.224\times$& $2.688\times$& $3.316\times$& $5.254\times$\\
\textbf{MobileNet V2} &Image Classification & Accuracy & 71.62& 71.44&$8.380\times$ & $2.073\times$ & $3.169\times$ & $4.818\times$ & $6.030\times$&$71.35$ &$7.320\times$ & $1.252\times$ & $2.390\times$ & $3.358\times$ & $4.931\times$\\
\textbf{DLRM}           &Recommendation & Accuracy & 79.42 & 79.39& $14.58\times$& $2.123\times$& $2.724\times$& $5.334\times$& $6.191\times$ &79.45 & $7.041\times$& $1.167\times$& $2.563\times$& $4.113\times$& $5.011\times$\\
\textbf{ViT}            &Image Modeling Pre-training & Evaluation Loss & 0.087& 0.087& $313.5\times$&$5.947\times$ & $10.84\times$ & $13.23\times$& $17.66\times$&0.092& $151.68\times$&$3.095\times$ & $7.426\times$ & $8.909\times$& $9.741\times$\\
\textbf{GPT-2}          &Language Modeling Fine-tuning & Perplexity & 20.95 & 21.13& $5.506\times$&$1.822\times$ & $2.357\times$ & $3.345\times$& $3.734\times$&21.12& $4.159\times$&$1.065\times$ & $2.165\times$ & $2.454\times$& $3.469\times$\\
\textbf{BERT - CoLA}           &Text Classification Fine-tuning - CoLA & Matthews Correlation & 55.99 & 57.03& $9.878\times$&$1.996\times$ & $2.680\times$ & $4.362\times$& $5.069\times$&56.11& $6.864\times$&$1.034\times$ & $2.085\times$ & $2.886\times$& $4.452\times$\\
\textbf{BERT - SST-2}           &Text Classification Fine-tuning - SST-2 & Accuracy & 93.23 & 91.97& $15.31\times$&$2.114\times$ & $2.806\times$ & $5.090\times$& $5.976\times$&92.44& $6.180\times$&$1.032\times$ & $2.080\times$ & $2.789\times$& $4.228\times$\\
\textbf{BERT - MRPC}           &Text Classification Fine-tuning - MRPC & Accuracy & 84.56 &84.80 & $9.189\times$&$1.988\times$ & $2.601\times$ & $4.252\times$& $4.864\times$&84.59& $3.574\times$&$1.016\times$ & $2.082\times$ & $2.237\times$& $3.114\times$\\
\textbf{BERT - STS-B}           &Text Classification Fine-tuning - STS-B & Pearson & 88.92 & 88.81& $6.322\times$&$1.821\times$ & $2.465\times$ & $3.544\times$& $4.059\times$&89.11& $3.501\times$&$1.051\times$ & $2.078\times$ & $2.257\times$& $3.071\times$\\
\textbf{BERT - QQP }           &Text Classification Fine-tuning - QQP & Accuracy & 90.71 & 90.30& $11.69\times$&$1.970\times$ & $2.585\times$ & $4.553\times$& $5.278\times$&90.43& $6.689\times$&$1.119\times$ & $2.073\times$ & $3.022\times$& $4.385\times$\\
\textbf{BERT - MNLI }           &Text Classification Fine-tuning - MNLI & Matched Accuracy & 83.87 & 84.16& $9.786\times$&$1.884\times$ & $2.470\times$ & $4.213\times$& $4.856\times$&84.03& $6.732\times$&$1.069\times$ & $2.073\times$ & $2.936\times$& $4.398\times$\\
\textbf{BERT - QNLI }           &Text Classification Fine-tuning - QNLI & Accuracy & 90.54 & 90.28& $9.342\times$&$1.903\times$ & $2.486\times$ & $4.175\times$& $4.791\times$&90.54& $6.552\times$&$1.503\times$ & $2.071\times$ & $3.005\times$& $4.340\times$\\ \hline
\textbf{Geo Mean}           & &  &  & & $11.94\times$&$2.126\times$ & $2.972\times$ & $4.736\times$& $5.637\times$&& $7.556\times$&$1.228\times$ & $2.492\times$ & $3.185\times$& $4.558\times$
 
\end{tabular}
}

\end{table*}

\subsection{Exponent: \EC (\Gshort)}
\label{sec:exponent}
Exponents are biased 8b integers under the BFloat16 and FP32 formats, and even narrower when optimized with \QEshort or \BWEshort. Despite this, all exponents per tensor are recorded using the same bit count. During training, these values tend to cluster heavily around a number, as shown in Figure ~\ref{fig:original_distribution}, which depicts the exponent distribution for ResNet18 after the 10th epoch. Leveraging this skewed distribution, we apply a variable-length, \textit{lossless} encoding that adjusts bit usage to the actual value of each exponent, such as using only 2 bits for the value 3. This method involves subtracting a bias value from each exponent, allocating fewer bits to frequent values and more to rare ones, thus minimizing average bit usage. A 3b metadata field records the bitlength, shared across multiple exponents to minimize metadata overhead. We also observe that these values are spatially correlated, with proximal values often being similar.

\EC encoding operates as follows: Given a tensor, \EC groups values into sets of 8. Exponents are encoded as $E - \mathit{bias}$, where $E$ is the original exponent and the bias is a fixed value; our experiments show that using 127 as the bias provides the best compression ratio. A leading 1 detector finds how many bits are needed for the largest exponent. That bitlength is recorded as metadata using 3 bits. All exponents of the group are stored using this bitlength. Using variable bitlengths for encoding values complicates random access, as it precludes direct computation of addresses in the tensor using their indices. However, deep learning workloads typically do not require random access to DRAM. Instead, blocking for data reuse leads to long sequential accesses to DRAM, which are conducive to the use of variable length containers. Variable bitlength encoding of values is common in quantization and memory compression methods~\cite{Deep_Compression,han_eie:isca_2016,ShapeshifterMICRO}.

\noindent\textbf{Evaluation: Bitlength: }
We measure the number of bits needed for the exponents using \EC during the training of ResNet18. Figure~\ref{fig:compressed_distribution} reports the cumulative distributions of exponent bitlength for a batch across: 1)~all layers and 2)~a single layer, for weights and activations. After encoding, more than 90\% of the weight exponents require 4b or fewer, while almost 90\% of activation exponents require 2b or fewer. Across training, the compression ratios for weight and activation exponents are $0.60$ and $0.38$, respectively.

\section{Evaluation}
\label{sct:evaluation}
We study the effects of \QMQEshort and \BWMBWEshort with, and without \EC(\Gshort). We \textit{fully} train ResNet18, ResNet50 and MobileNet V2 on ImageNet, DLRM on Kaggle Criteo as well as pre-train ViT on Cifar10, finetune BERT on GLUE and GPT-2 on Wikitext 2, using an RTX3090/24GB with PyTorch v1.10. We implement \QMQEshort by modifying the loss function and adding the gradient calculations for the per tensor parameters. We simulate \BWMBWEshort in software. For both methods, we \textit{faithfully} emulate \textit{all} bitlength arithmetic effects by truncating the mantissa bits and encoding/decoding exponents at the boundary of each layer using PyTorch hooks and custom layers. We measure \EC's effects in software via hooks. These changes allow us to measure the effects our methods have on traffic and accuracy. The following discussion is illustrated for ResNet18 in Figure~\ref{fig:resnet18_mem_reduction} and shown in full detail for all networks in Table~\ref{tbl:total_mem_reduction}.

\subsection{Memory Footprint Reduction}
\label{sec:comb}

First, we report cumulative memory footprint reduction and validation accuracy in comparison with FP32 in Table~\ref{tbl:total_mem_reduction}. Our compression techniques excel at reducing footprint, with little effect on accuracy. 

\QMQEshort reduces the total training footprint by $3.35\times$ to $13.23\times$ with an average of $4.74\times$. While \QMQEshort works great on exponent as well, it is exceptionally good at compressing mantissas. With the addition of \Gshort, the benefits further extend to $3.73\times$ to $17.66\times$ with an average of $5.64\times$. \BWMBWEshort on the other hand reduces the total training footprint by $2.24\times$ to $8.91\times$ with an average of $3.19\times$ without, and $3.07\times$ to $9.74\times$ with an average of $4.56\times$ with \Gshort, respectively. While \BWMBWEshort provides great compression rate for mantissas, it is less effective for exponents. The addition of \Gshort recovers most of the compression gap. In the end, \QMQEshort outperforms \BWMBWEshort in every single case. However, this comes with the \textit{slightly} larger overhead.

The optional \Gshort boosts the compression rate by $19\%$ and $43\%$ on top of \QMQEshort and \BWMBWEshort, respectively. It works far better for \BWMBWEshort because the method greatly removes outliers by focusing on exponent range, and helps it recover almost all of the exponent compression gap. This comes at the cost of variable tensor sizes, and therefore inability of random memory accesses to the off-chip memory. Fortunately, training only requires sequential access to off-chip memory, and sequential/strided/random accesses to on-chip memory which are fully supported by our design.

\subsection{Quantization Alternatives}

Alternative quantization approaches require selecting a datatype for training and sticking to it. The choice practically boils down to FP32, Bfloat16, and FP8. Table~\ref{tbl:total_mem_reduction} shows memory reduction in comparison with FP32. Assuming that the network converges with the smaller datatype, Bfloat16 would always reduce the footprint by $2\times$ and FP8 by $4\times$. Every single combination of our methods/networks outperforms FP32 and BFloat16 by a significant margin. 

Furthermore, \QMQEshort produces a $16\%$ smaller footprint than FP8 with GPT-2 being the only network where FP8 wins. With \Gshort, \QMQEshort's advantage increases to $29\%$.

The case for \BWMBWEshort vs FP8 isn't as clear cut, outperforming FP8 in all non CNNs. With \Gshort, \BWMBWEshort, on average, produces a $12\%$ smaller footprint than FP8.

Another key benefit of our methods is that they are \textit{adaptable}. Choosing FP8 is risky, since the results of training is only evident at the end. \cite{fp8_Nvidia_Intel_Arm} show that FP8 is a good choice for many networks, but they also note that there are architectures for which it is not sufficient. Our method provides a greater certainty of success, whilst obtaining better footprint.

\begin{table}
\centering
\caption{\BP: ResNet18 Validation Accuracy and total memory reduction vs. FP32.}
\label{tbl:quantum_intiger}
\resizebox{0.85\columnwidth}{!}
{
\begin{tabular}{l|l|ll}
\textbf{}                  & \textbf{FP32} & \multicolumn{2}{c}{\textbf{\BP}} \\
\textbf{Network} & \textbf{Accuracy}& \textbf{Accuracy} & \textbf{Footprint Reduction}\\ \hline
\textbf{ResNet18}       & 69.94 & 69.15 & $6.21\times$ \\
\end{tabular}
}
\end{table}

\begin{figure}[t]
\centering
\subfloat[Performance]{
\includegraphics[width=0.95\columnwidth]{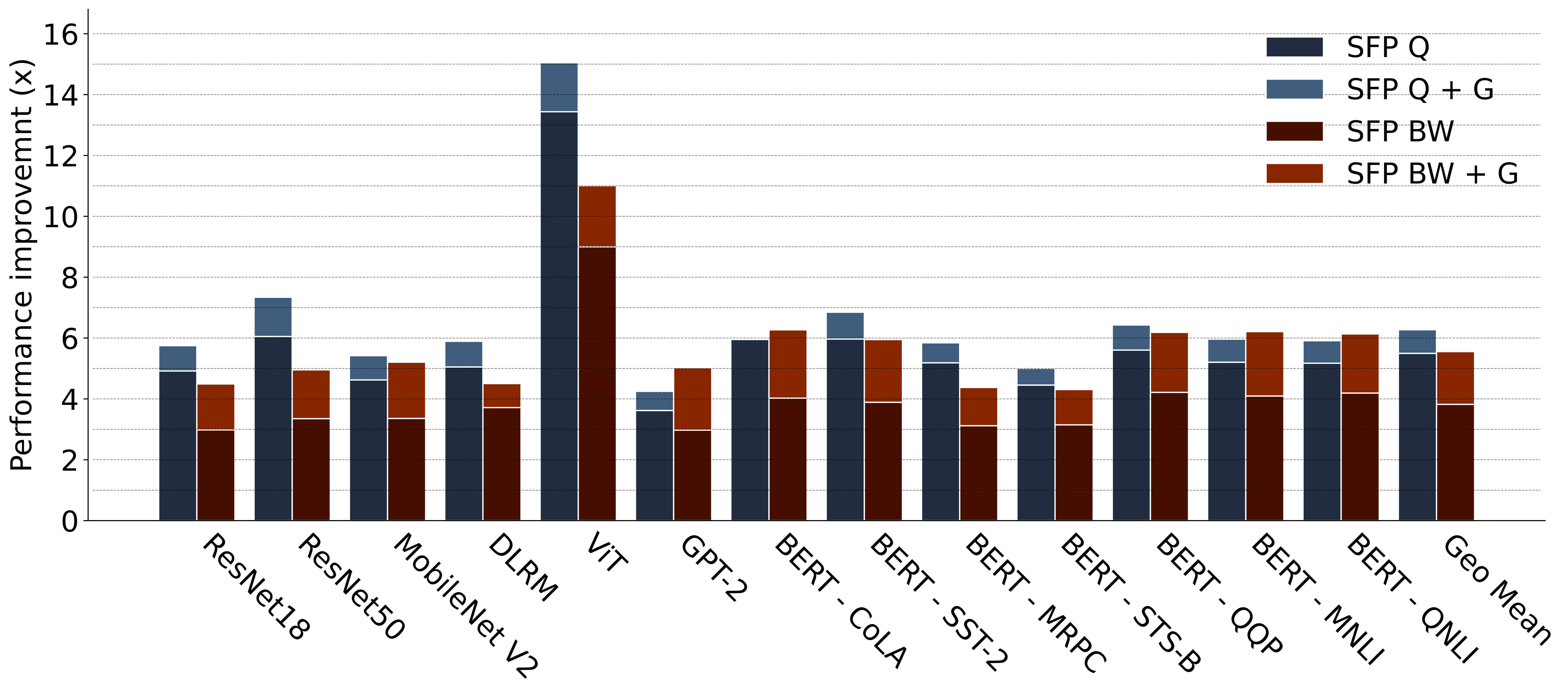}
\label{fig:32bFP_perf_comparison}
}

\subfloat[Energy Efficiency]{
\includegraphics[width=0.95\columnwidth]{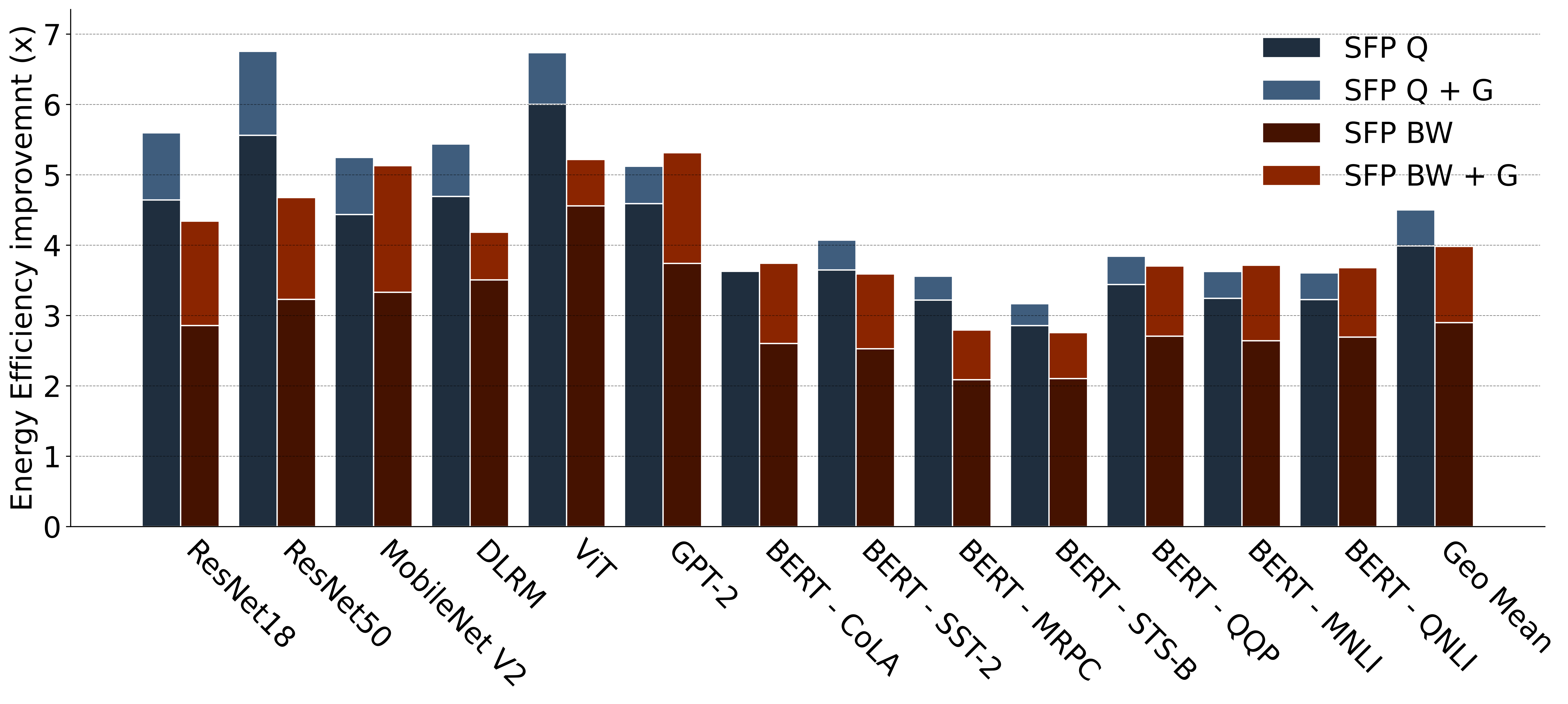}
\label{fig:32bFP_ener_comparison}
}
\caption{Comparison with FP32.}
\label{fig:32bFP_simulation}
\end{figure}

\begin{figure}[t]
\centering
\subfloat[Performance]{
\includegraphics[width=0.95\columnwidth]{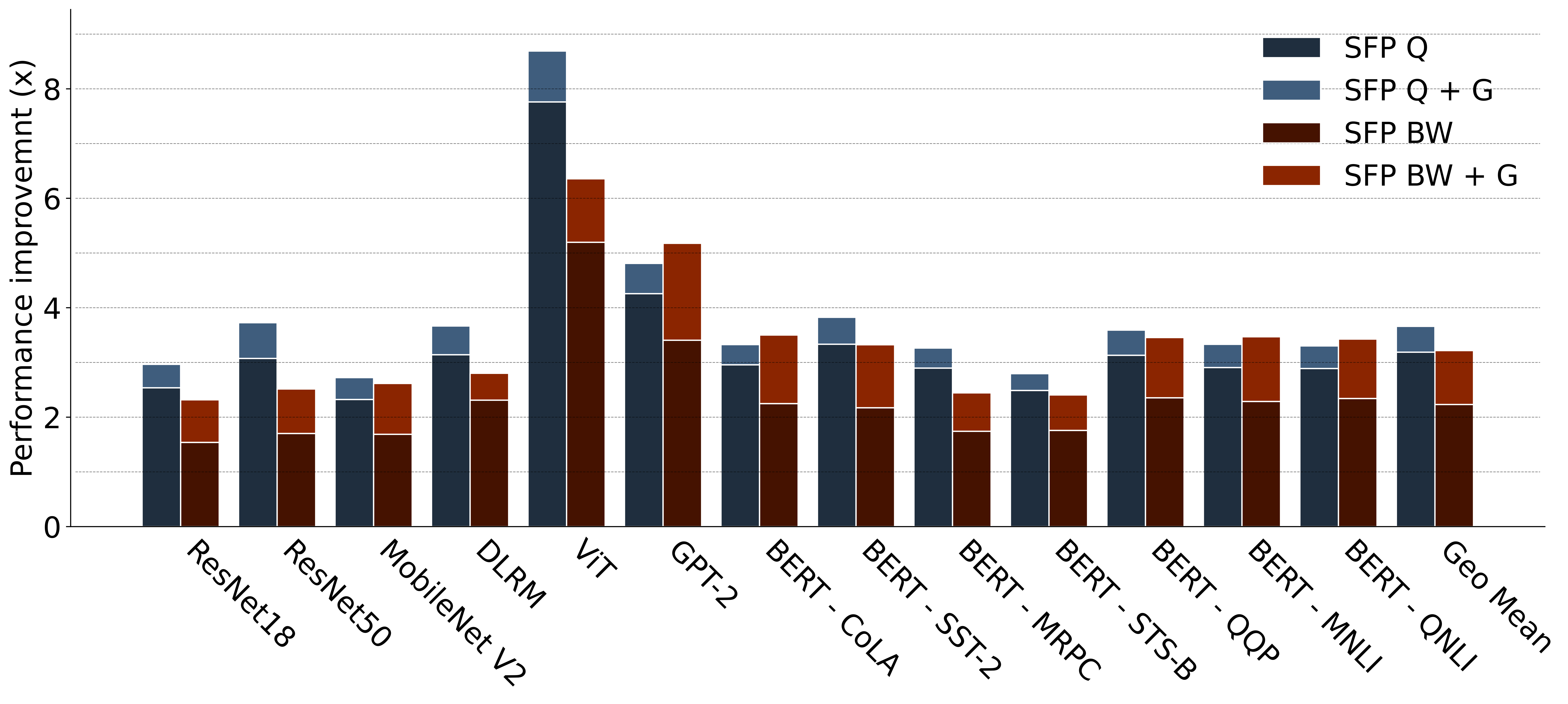}
\label{fig:16bFP_perf_comparison}
}

\subfloat[Energy Efficiency]{
\includegraphics[width=0.95\columnwidth]{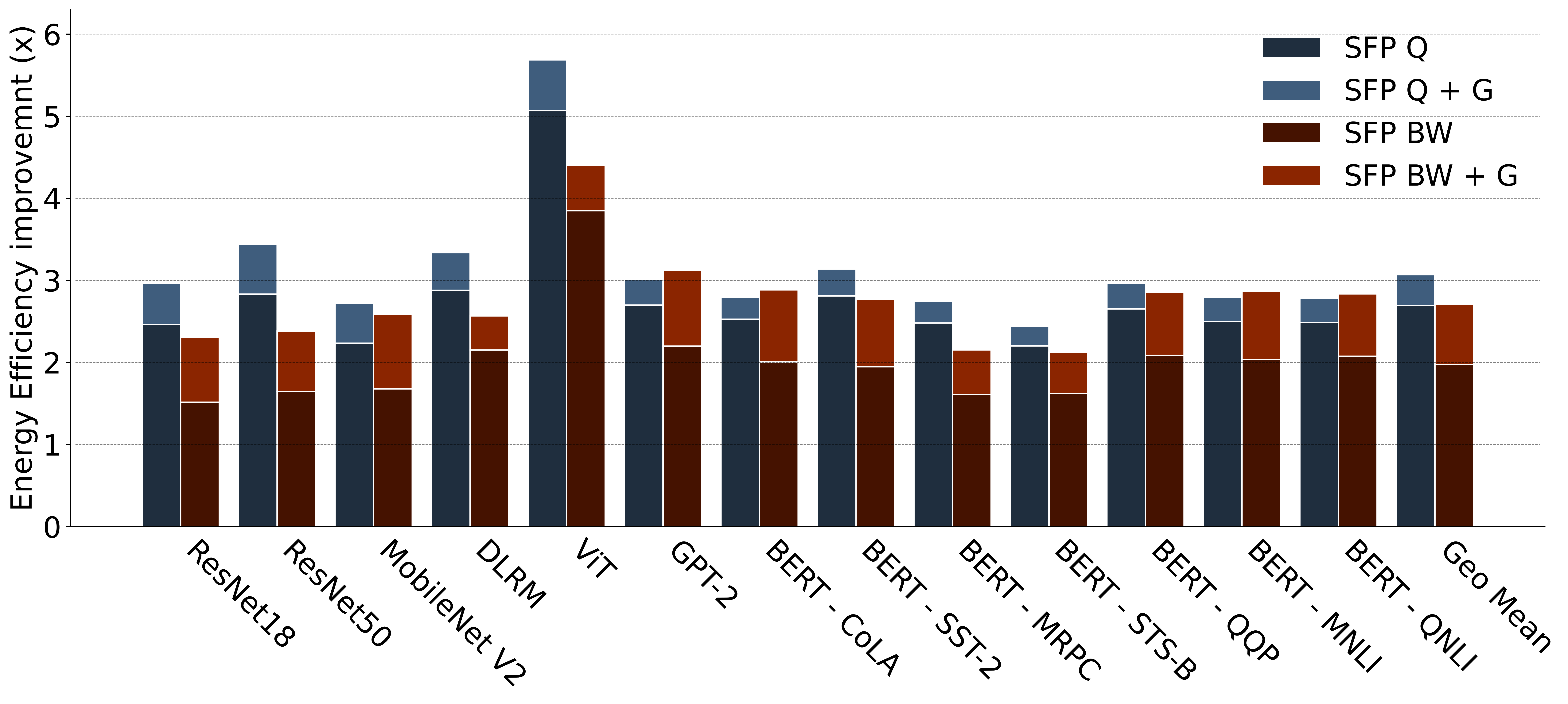}
\label{fig:16bFP_ener_comparison}
}
\caption{Comparison with BFloat16.}
\label{fig:16bFP_simulation}
\end{figure}

\begin{figure}[t]
\centering
\subfloat[Performance]{
\includegraphics[width=0.95\columnwidth]{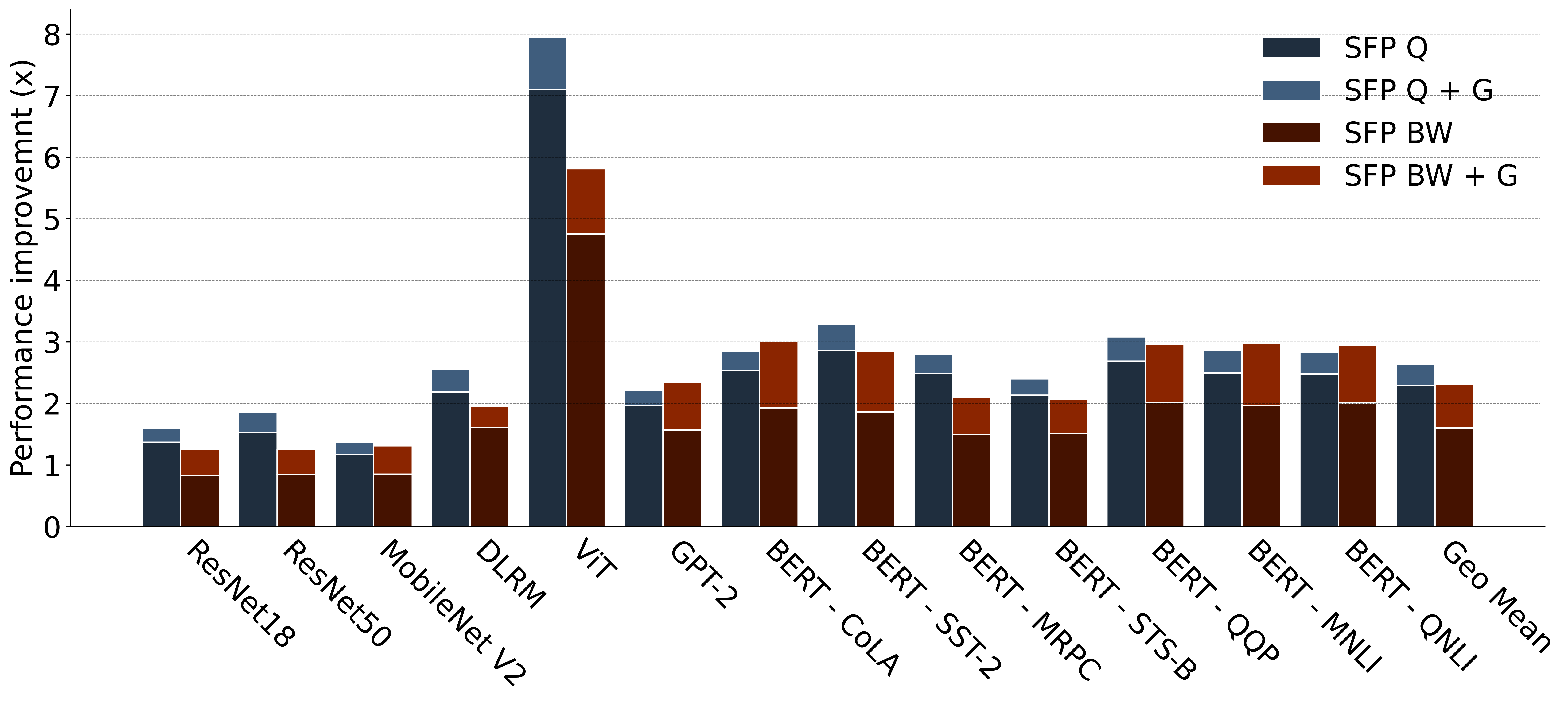}
\label{fig:8bFP_perf_comparison}
}

\subfloat[Energy Efficiency]{
\includegraphics[width=0.95\columnwidth]{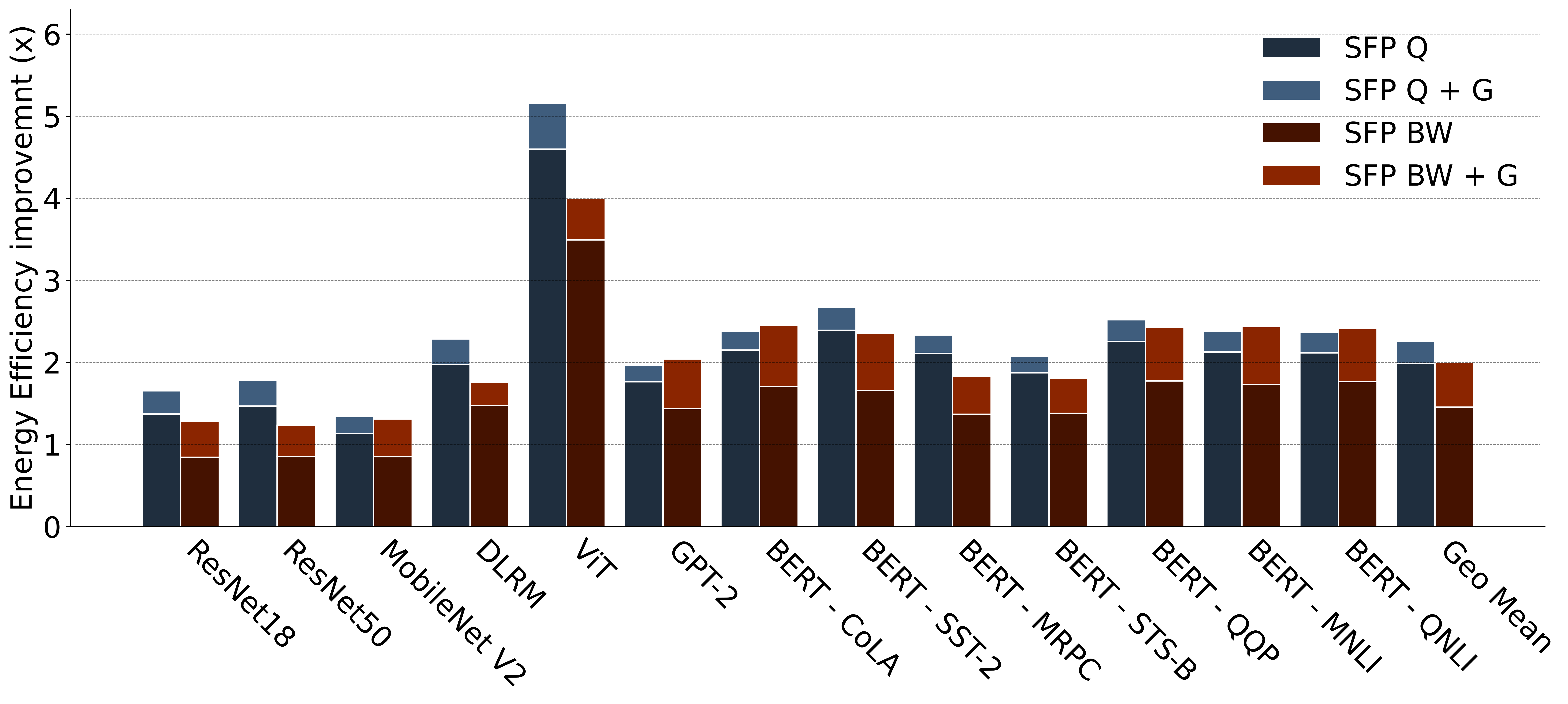}
\label{fig:8bFP_ener_comparison}
}
\caption{Comparison with FP8.}
\label{fig:8bFP_simulation}
\end{figure}

\subsection{\BP (\QIshort): Fixed-Point Datatype}

For some models, fixed-point training is possible. While our main goal is to learn the optimal \textit{floating-point} datatypes, \QMshort can easily be adapted to learn optimal fixed-point datatypes. One common way to train for fixed-point inference is by representing the activations in fixed point during training and to use integer arithmetic during the forward pass. The only modification we need is to switch out Eq.~\ref{eq:mantissa} for one that represents fixed-point. Other aspects of \QMshort stay the same, while exponent is not used. 

We present the footprint reduction and accuracy effect of the resulting \QIshort by showing results with ResNet18 on ImageNet in Table~\ref{tbl:quantum_intiger}. This simple, yet effective, modification learns the per-tensor optimal bitlengths for uniform quantization training with minimal accuracy cost. This is also a good choice for training when we are confident that the task the network is solving can be done in low bitlength fixed-point. The \QIshort behavior is very similar to \QMshort.

\section{Performance and Energy Efficiency}\label{sec:eval_perf_ee}
We evaluate execution time and energy efficiency with \DPRL for all networks listed in Table~\ref{tbl:total_mem_reduction}. Figures~\ref{fig:32bFP_simulation}, ~\ref{fig:16bFP_simulation} and ~\ref{fig:8bFP_simulation} report execution time and energy improvements in comparison with FP32, BFloat16 and FP8 baselines, respectively. The baseline datatype is used for weight updates and gradients. We further assume that all the networks can be trained to baseline accuracy with FP8 and BFloat16. However, this is not given.

\subsection{Hardware Evaluation Methodology}
\label{sec:hardware_evaluation_methodology}
We assess the execution time and energy efficiency by integrating \DPRL units into a hardware accelerator that reflects state-of-the-art designs. The accelerator has 8k units (each capable of performing 4 MACs per cycle on the baseline datatype), and a 500MHz clock for a peak compute bandwidth of 16TFLOPS. It uses $8$ channels of LPDDR4-3200 DRAM memory and 32MB of on-chip buffers. 

Our evaluation relies on an analytical model of the accelerator. We use CACTI~\cite{Muralimanohar_cacti6.0} to model on-chip structures and DRAMSIM3~\cite{Dramsim3} to estimate the time and energy for off-chip memory accesses. The hardware units are implemented in Verilog, synthesized using the Synopsys Design Compiler~\cite{synopsysDC}, and are laid out with Cadence Innovus~\cite{innovus}. The area overhead for the compressor/decompressor proves negligible, taking only $0.36\%$ of the total accelerator area, excluding on-chip memory. 

Power estimation is performed in Innovus using traces from a representative sample to accurately model signal activity. Appendix~\ref{sec:hardware} describes our hardware units and the accelerator. Appendix~\ref{sec:eval_method} explains in more detail the evaluation methodology and the analytical model used for execution time and energy efficiency estimation.

\subsection{Hardware Evaluation}
We first compare \DPRL with the FP32 and BFloat16 baselines. FP32 has been the safe choice for training. Similarly, BFloat16 is another common choice for training. Figures~\ref{fig:32bFP_simulation} and ~\ref{fig:16bFP_simulation} show that all versions of \DPRL greatly outperform both the FP32 and BFloat16 baselines in performance and energy efficiency, for every single network.

FP8 is a riskier datatype for training. For some networks and tasks it is sufficient, but for others it is not. Figure~\ref{fig:8bFP_simulation} shows that both \QMQEshort and \QMQEGshort noticeably outperform FP8 in both performance and energy efficiency, for all networks. The figure also shows that \BWMBWEshort generally outperforms FP8, having a better average performance and energy efficiency. The outliers where results would be better with FP8 are all for ImageNet CNNs. MobileNet V2, one of the three outliers, can not be trained in FP8 because doing so introduces a noticeable accuracy loss. However, \BWMBWEshort can be used as shown in Table~\ref{tbl:total_mem_reduction}. Finally, our methods coupled with \EC outperform FP8 on every network, both in performance and energy efficiency.

In general, as the baseline gets smaller, training becomes riskier, and the margins by which all \DPRL methods outperform become smaller. At the boundary where the accuracy starts to degrade, our methods are still significantly faster and more energy efficient. 

\QMQEshort outperforms \BWMBWEshort across all networks. However, when used with \EC the difference between our two methods is not as pronounced. \QMQEGshort is still better but the difference is smaller and in some cases it even reverses (e.g. GPT-2).

Finally, we assumed that all benefits from smaller datatypes come from off-chip memory transfers. For accelerators where compute can also be made more efficient through using smaller, spatially composable or bit serial compute units, improvement with our methods would be even greater.

\section{Conclusion}

We introduced methods that dynamically adapt the bitlengths and containers used for floating-point values during training. The different distributions of the exponents and mantissas led us to tailored approaches for each. We target the largest contributors to off-chip traffic during training for both activations and weights. In addition, in the case where fixed-point training is preferred, we showed the effectiveness of our approach to determine the best containers used for fixed-point values during training. 
To our knowledge, this is the first work that demonstrates how to:
(1) \textit{determine} and (2) \textit{continuously adjust} the memory containers (how many bits should be used when storing floating-point mantissas and exponents in memory), and to do so (3) \textit{on-the-fly}, for the purpose of (4) making \textit{training} itself faster and/or more energy efficient.  
There are several directions for improvements and further exploration including expanding the methods to also target the gradients and refining the underlying policies they use to adapt mantissa lengths. Regardless, this work has demonstrated that the methods are effective and superior to using fixed preselected datatypes. The key advantages of our methods are: 1)~they are dynamic and adaptive, 2)~they do not modify the training algorithm, 3) they will naturally extend to future algorithms without modifications and 4)~they take advantage of value content.

\bibliography{example_paper}

\begin{thebibliography}{10}
\providecommand{\url}[1]{#1}
\csname url@samestyle\endcsname
\providecommand{\newblock}{\relax}
\providecommand{\bibinfo}[2]{#2}
\providecommand{\BIBentrySTDinterwordspacing}{\spaceskip=0pt\relax}
\providecommand{\BIBentryALTinterwordstretchfactor}{4}
\providecommand{\BIBentryALTinterwordspacing}{\spaceskip=\fontdimen2\font plus
\BIBentryALTinterwordstretchfactor\fontdimen3\font minus \fontdimen4\font\relax}
\providecommand{\BIBforeignlanguage}[2]{{%
\expandafter\ifx\csname l@#1\endcsname\relax
\typeout{** WARNING: IEEEtran.bst: No hyphenation pattern has been}%
\typeout{** loaded for the language `#1'. Using the pattern for}%
\typeout{** the default language instead.}%
\else
\language=\csname l@#1\endcsname
\fi
#2}}
\providecommand{\BIBdecl}{\relax}
\BIBdecl

\bibitem{GIST}
\BIBentryALTinterwordspacing
A.~Jain, A.~Phanishayee, J.~Mars, L.~Tang, and G.~Pekhimenko, ``Gist: Efficient data encoding for deep neural network training,'' in \emph{Proceedings of the 45th Annual International Symposium on Computer Architecture}, ser. ISCA '18.\hskip 1em plus 0.5em minus 0.4em\relax Piscataway, NJ, USA: IEEE Press, 2018, pp. 776--789. [Online]. Available: \url{https://doi.org/10.1109/ISCA.2018.00070}
\BIBentrySTDinterwordspacing

\bibitem{HorowitzEnergy}
M.~Horowitz, ``1.1 computing's energy problem (and what we can do about it),'' in \emph{2014 IEEE International Solid-State Circuits Conference Digest of Technical Papers (ISSCC)}, 2014, pp. 10--14.

\bibitem{bfloat16}
\BIBentryALTinterwordspacing
D.~D. Kalamkar, D.~Mudigere, N.~Mellempudi, D.~Das, K.~Banerjee, S.~Avancha, D.~T. Vooturi, N.~Jammalamadaka, J.~Huang, H.~Yuen, J.~Yang, J.~Park, A.~Heinecke, E.~Georganas, S.~Srinivasan, A.~Kundu, M.~Smelyanskiy, B.~Kaul, and P.~Dubey, ``A study of {BFLOAT16} for deep learning training,'' \emph{CoRR}, vol. abs/1905.12322, 2019. [Online]. Available: \url{http://arxiv.org/abs/1905.12322}
\BIBentrySTDinterwordspacing

\bibitem{DBLP:conf/iclr/0002MMKAB0VKGHD18}
\BIBentryALTinterwordspacing
D.~Das, N.~Mellempudi, D.~Mudigere, D.~D. Kalamkar, S.~Avancha, K.~Banerjee, S.~Sridharan, K.~Vaidyanathan, B.~Kaul, E.~Georganas, A.~Heinecke, P.~Dubey, J.~Corbal, N.~Shustrov, R.~Dubtsov, E.~Fomenko, and V.~O. Pirogov, ``Mixed precision training of convolutional neural networks using integer operations,'' in \emph{6th International Conference on Learning Representations, {ICLR} 2018, Vancouver, BC, Canada, April 30 - May 3, 2018, Conference Track Proceedings}, 2018. [Online]. Available: \url{https://openreview.net/forum?id=H135uzZ0-}
\BIBentrySTDinterwordspacing

\bibitem{Koster:2017:FAN:3294771.3294937}
\BIBentryALTinterwordspacing
U.~K\"{o}ster, T.~J. Webb, X.~Wang, M.~Nassar, A.~K. Bansal, W.~H. Constable, O.~H. Elibol, S.~Gray, S.~Hall, L.~Hornof, A.~Khosrowshahi, C.~Kloss, R.~J. Pai, and N.~Rao, ``Flexpoint: An adaptive numerical format for efficient training of deep neural networks,'' in \emph{Proceedings of the 31st International Conference on Neural Information Processing Systems}, ser. NIPS'17.\hskip 1em plus 0.5em minus 0.4em\relax USA: Curran Associates Inc., 2017, pp. 1740--1750. [Online]. Available: \url{http://dl.acm.org/citation.cfm?id=3294771.3294937}
\BIBentrySTDinterwordspacing

\bibitem{mixedP}
\BIBentryALTinterwordspacing
P.~Micikevicius, S.~Narang, J.~Alben, G.~F. Diamos, E.~Elsen, D.~Garc{\'{\i}}a, B.~Ginsburg, M.~Houston, O.~Kuchaiev, G.~Venkatesh, and H.~Wu, ``Mixed precision training,'' in \emph{6th International Conference on Learning Representations, {ICLR} 2018, Vancouver, BC, Canada, April 30 - May 3, 2018, Conference Track Proceedings}, 2018. [Online]. Available: \url{https://openreview.net/forum?id=r1gs9JgRZ}
\BIBentrySTDinterwordspacing

\bibitem{nvidia_mixedP}
NVIDIA, ``Training with mixed precision,'' \url{https://docs.nvidia.com/deeplearning/sdk/mixed-precision-training/index.html}.

\bibitem{Drumond:2018:TDH:3326943.3326985}
\BIBentryALTinterwordspacing
M.~Drumond, T.~Lin, M.~Jaggi, and B.~Falsafi, ``Training {DNNs} with hybrid block floating point,'' in \emph{Proceedings of the 32Nd International Conference on Neural Information Processing Systems}, ser. NIPS'18.\hskip 1em plus 0.5em minus 0.4em\relax USA: Curran Associates Inc., 2018, pp. 451--461. [Online]. Available: \url{http://dl.acm.org/citation.cfm?id=3326943.3326985}
\BIBentrySTDinterwordspacing

\bibitem{IBM_8bit}
N.~Wang, J.~Choi, D.~Brand, C.-Y. Chen, and K.~Gopalakrishnan, ``Training deep neural networks with 8-bit floating point numbers,'' in \emph{Proceedings of the 32nd International Conference on Neural Information Processing Systems}, ser. NIPS'18.\hskip 1em plus 0.5em minus 0.4em\relax Red Hook, NY, USA: Curran Associates Inc., 2018, pp. 7686--7695.

\bibitem{IBM_4bit}
\BIBentryALTinterwordspacing
X.~Sun, N.~Wang, C.-Y. Chen, J.~Ni, A.~Agrawal, X.~Cui, S.~Venkataramani, K.~El~Maghraoui, V.~V. Srinivasan, and K.~Gopalakrishnan, ``Ultra-low precision 4-bit training of deep neural networks,'' in \emph{Advances in Neural Information Processing Systems}, H.~Larochelle, M.~Ranzato, R.~Hadsell, M.~Balcan, and H.~Lin, Eds., vol.~33.\hskip 1em plus 0.5em minus 0.4em\relax Curran Associates, Inc., 2020, pp. 1796--1807. [Online]. Available: \url{https://proceedings.neurips.cc/paper/2020/file/13b919438259814cd5be8cb45877d577-Paper.pdf}
\BIBentrySTDinterwordspacing

\bibitem{fp8_Nvidia_Intel_Arm}
P.~Micikevicius, D.~Stosic, N.~Burgess, M.~Cornea, P.~Dubey, R.~Grisenthwaite, S.~Ha, A.~Heinecke, P.~Judd, J.~Kamalu \emph{et~al.}, ``Fp8 formats for deep learning,'' \emph{arXiv preprint arXiv:2209.05433}, 2022.

\bibitem{rouhani2023microscaling}
B.~D. Rouhani, R.~Zhao, A.~More, M.~Hall, A.~Khodamoradi, S.~Deng, D.~Choudhary, M.~Cornea, E.~Dellinger, K.~Denolf, S.~Dusan, V.~Elango, M.~Golub, A.~Heinecke, P.~James-Roxby, D.~Jani, G.~Kolhe, M.~Langhammer, A.~Li, L.~Melnick, M.~Mesmakhosroshahi, A.~Rodriguez, M.~Schulte, R.~Shafipour, L.~Shao, M.~Siu, P.~Dubey, P.~Micikevicius, M.~Naumov, C.~Verrilli, R.~Wittig, D.~Burger, and E.~Chung, ``Microscaling data formats for deep learning,'' 2023.

\bibitem{FPRaker}
\BIBentryALTinterwordspacing
O.~M. Awad, M.~Mahmoud, I.~Edo, A.~H. Zadeh, C.~Bannon, A.~Jayarajan, G.~Pekhimenko, and A.~Moshovos, ``Fpraker: {A} processing element for accelerating neural network training,'' in \emph{{MICRO} '21: 54th Annual {IEEE/ACM} International Symposium on Microarchitecture, Virtual Event, Greece, October 18-22, 2021}.\hskip 1em plus 0.5em minus 0.4em\relax {ACM}, 2021, pp. 857--869. [Online]. Available: \url{https://doi.org/10.1145/3466752.3480106}
\BIBentrySTDinterwordspacing

\bibitem{Deep_Compression}
\BIBentryALTinterwordspacing
S.~Han, H.~Mao, and W.~J. Dally, ``Deep compression: Compressing deep neural network with pruning, trained quantization and huffman coding,'' in \emph{4th International Conference on Learning Representations, {ICLR} 2016, San Juan, Puerto Rico, May 2-4, 2016, Conference Track Proceedings}, Y.~Bengio and Y.~LeCun, Eds., 2016. [Online]. Available: \url{http://arxiv.org/abs/1510.00149}
\BIBentrySTDinterwordspacing

\bibitem{ShapeshifterMICRO}
\BIBentryALTinterwordspacing
A.~D. Lascorz, S.~Sharify, I.~Edo, D.~M. Stuart, O.~M. Awad, P.~Judd, M.~Mahmoud, M.~Nikoli\'{c}, K.~Siu, Z.~Poulos, and A.~Moshovos, ``Shapeshifter: Enabling fine-grain data width adaptation in deep learning,'' in \emph{Proceedings of the 52nd Annual IEEE/ACM International Symposium on Microarchitecture}, ser. MICRO 52.\hskip 1em plus 0.5em minus 0.4em\relax New York, NY, USA: Association for Computing Machinery, 2019. [Online]. Available: \url{https://doi.org/10.1145/3352460.3358295}
\BIBentrySTDinterwordspacing

\bibitem{han_eie:isca_2016}
S.~Han, X.~Liu, H.~Mao, J.~Pu, A.~Pedram, M.~A. Horowitz, and W.~J. Dally, ``Eie: Efficient inference engine on compressed deep neural network,'' in \emph{Intl' Symp. on Computer Architecture}, 2016.

\bibitem{MX_block_floating_point}
B.~Rouhani, R.~Zhao, V.~Elango, R.~Shafipour, M.~Hall, M.~Mesmakhosroshahi, A.~More, L.~Melnick, M.~Golub, G.~Varatkar, L.~Shao, G.~Kolhe, D.~Melts, J.~Klar, R.~L'Heureux, M.~Perry, D.~Burger, E.~Chung, Z.~Deng, S.~Naghshineh, J.~Park, and M.~Naumov, ``With shared microexponents, a little shifting goes a long way,'' 2023.

\bibitem{FASTDNN}
S.~Qian~Zhang, B.~McDanel, and H.~T. Kung, ``Fast: Dnn training under variable precision block floating point with stochastic rounding,'' in \emph{2022 IEEE International Symposium on High-Performance Computer Architecture (HPCA)}, 2022, pp. 846--860.

\bibitem{CVPR_FixedPointBackPropagationTraining}
X.~Zhang, S.~Liu, R.~Zhang, C.~Liu, D.~Huang, S.~Zhou, J.~Guo, Q.~Guo, Z.~Du, T.~Zhi, and Y.~Chen, ``Fixed-point back-propagation training,'' in \emph{2020 IEEE/CVF Conference on Computer Vision and Pattern Recognition (CVPR)}, 2020, pp. 2327--2335.

\bibitem{evans2021acgc}
\BIBentryALTinterwordspacing
R.~D. Evans and T.~Aamodt, ``{AC}-{GC}: Lossy activation compression with guaranteed convergence,'' in \emph{Advances in Neural Information Processing Systems}, A.~Beygelzimer, Y.~Dauphin, P.~Liang, and J.~W. Vaughan, Eds., 2021. [Online]. Available: \url{https://openreview.net/forum?id=MwFdqFRxIF0}
\BIBentrySTDinterwordspacing

\bibitem{HAQ}
\BIBentryALTinterwordspacing
K.~Wang, Z.~Liu, Y.~Lin, J.~Lin, and S.~Han, ``{HAQ:} hardware-aware automated quantization,'' \emph{CoRR}, vol. abs/1811.08886, 2018. [Online]. Available: \url{http://arxiv.org/abs/1811.08886}
\BIBentrySTDinterwordspacing

\bibitem{bitpruning}
M.~Nikoli\'{c}, G.~B. Hacene, C.~Bannon, A.~D. Lascorz, M.~Courbariaux, Y.~Bengio, V.~Gripon, and A.~Moshovos, ``Bitpruning: Learning bitlengths for aggressive and accurate quantization,'' 2020.

\bibitem{SDQ}
\BIBentryALTinterwordspacing
X.~Huang, Z.~Shen, S.~Li, Z.~Liu, H.~Xianghong, J.~Wicaksana, E.~Xing, and K.-T. Cheng, ``{SDQ}: Stochastic differentiable quantization with mixed precision,'' in \emph{Proceedings of the 39th International Conference on Machine Learning}, ser. Proceedings of Machine Learning Research, K.~Chaudhuri, S.~Jegelka, L.~Song, C.~Szepesvari, G.~Niu, and S.~Sabato, Eds., vol. 162.\hskip 1em plus 0.5em minus 0.4em\relax PMLR, 17--23 Jul 2022, pp. 9295--9309. [Online]. Available: \url{https://proceedings.mlr.press/v162/huang22h.html}
\BIBentrySTDinterwordspacing

\bibitem{DNAS_Quantization}
\BIBentryALTinterwordspacing
B.~Wu, Y.~Wang, P.~Zhang, Y.~Tian, P.~Vajda, and K.~Keutzer, ``Mixed precision quantization of convnets via differentiable neural architecture search,'' \emph{CoRR}, vol. abs/1812.00090, 2018. [Online]. Available: \url{http://arxiv.org/abs/1812.00090}
\BIBentrySTDinterwordspacing

\bibitem{yang2021bsq}
\BIBentryALTinterwordspacing
H.~Yang, L.~Duan, Y.~Chen, and H.~Li, ``{\{}BSQ{\}}: Exploring bit-level sparsity for mixed-precision neural network quantization,'' in \emph{International Conference on Learning Representations}, 2021. [Online]. Available: \url{https://openreview.net/forum?id=TiXl51SCNw8}
\BIBentrySTDinterwordspacing

\bibitem{DBLP:conf/iiswc/NikolicMM18}
M.~Nikoli\'{c}, M.~Mahmoud, and A.~Moshovos, ``Characterizing sources of ineffectual computations in deep learning networks,'' in \emph{2018 IEEE International Symposium on Workload Characterization (IISWC)}, 2018, pp. 86--87.

\bibitem{resnet}
\BIBentryALTinterwordspacing
K.~He, X.~Zhang, S.~Ren, and J.~Sun, ``Deep residual learning for image recognition,'' \emph{CoRR}, vol. abs/1512.03385, 2015. [Online]. Available: \url{http://arxiv.org/abs/1512.03385}
\BIBentrySTDinterwordspacing

\bibitem{sandler2018mobilenetv2}
M.~Sandler, A.~Howard, M.~Zhu, A.~Zhmoginov, and L.-C. Chen, ``{MobileNetV2:} inverted residuals and linear bottlenecks,'' in \emph{IEEE Conf. on Computer Vision and Pattern Recognition}, 2018.

\bibitem{imagenet}
O.~Russakovsky, J.~Deng, H.~Su, J.~Krause, S.~Satheesh, S.~Ma, Z.~Huang, A.~Karpathy, A.~Khosla, M.~Bernstein, A.~C. Berg, and L.~Fei-Fei, ``{ImageNet} {Large} {Scale} {Visual} {Recognition} {Challenge},'' \emph{arXiv:1409.0575 [cs]}, Sep. 2014, arXiv: 1409.0575.

\bibitem{DLRM19}
\BIBentryALTinterwordspacing
M.~Naumov, D.~Mudigere, H.~M. Shi, J.~Huang, N.~Sundaraman, J.~Park, X.~Wang, U.~Gupta, C.~Wu, A.~G. Azzolini, D.~Dzhulgakov, A.~Mallevich, I.~Cherniavskii, Y.~Lu, R.~Krishnamoorthi, A.~Yu, V.~Kondratenko, S.~Pereira, X.~Chen, W.~Chen, V.~Rao, B.~Jia, L.~Xiong, and M.~Smelyanskiy, ``Deep learning recommendation model for personalization and recommendation systems,'' \emph{CoRR}, vol. abs/1906.00091, 2019. [Online]. Available: \url{https://arxiv.org/abs/1906.00091}
\BIBentrySTDinterwordspacing

\bibitem{vit}
\BIBentryALTinterwordspacing
A.~Dosovitskiy, L.~Beyer, A.~Kolesnikov, D.~Weissenborn, X.~Zhai, T.~Unterthiner, M.~Dehghani, M.~Minderer, G.~Heigold, S.~Gelly, J.~Uszkoreit, and N.~Houlsby, ``An image is worth 16x16 words: Transformers for image recognition at scale,'' 2020. [Online]. Available: \url{https://arxiv.org/abs/2010.11929}
\BIBentrySTDinterwordspacing

\bibitem{cifar10}
A.~Krizhevsky, ``Learning multiple layers of features from tiny images,'' Tech. Rep., 2009.

\bibitem{BERT}
\BIBentryALTinterwordspacing
J.~Devlin, M.~Chang, K.~Lee, and K.~Toutanova, ``{BERT:} pre-training of deep bidirectional transformers for language understanding,'' \emph{CoRR}, vol. abs/1810.04805, 2018. [Online]. Available: \url{http://arxiv.org/abs/1810.04805}
\BIBentrySTDinterwordspacing

\bibitem{Glue}
A.~Wang, A.~Singh, J.~Michael, F.~Hill, O.~Levy, and S.~R. Bowman, ``{GLUE}: A multi-task benchmark and analysis platform for natural language understanding,'' 2019, in the Proceedings of ICLR.

\bibitem{GPT2}
A.~Radford, J.~Wu, R.~Child, D.~Luan, D.~Amodei, and I.~Sutskever, ``Language models are unsupervised multitask learners,'' 2019.

\bibitem{wikitext2}
\BIBentryALTinterwordspacing
S.~Merity, C.~Xiong, J.~Bradbury, and R.~Socher, ``Pointer sentinel mixture models,'' \emph{CoRR}, vol. abs/1609.07843, 2016. [Online]. Available: \url{http://arxiv.org/abs/1609.07843}
\BIBentrySTDinterwordspacing

\bibitem{Muralimanohar_cacti6.0}
N.~Muralimanohar and R.~Balasubramonian, ``Cacti 6.0: A tool to understand large caches.''

\bibitem{Dramsim3}
S.~Li, Z.~Yang, D.~Reddy, A.~Srivastava, and B.~Jacob, ``Dramsim3: A cycle-accurate, thermal-capable dram simulator,'' \emph{IEEE Computer Architecture Letters}, vol.~19, no.~2, pp. 106--109, 2020.

\bibitem{synopsysDC}
{S}ynopsys, ``{D}esign {C}ompiler,'' \url{http://www.synopsys.com/Tools/Implementation/RTLSynthesis/DesignCompiler/Pages}.

\bibitem{innovus}
Cadence, ``Innovus implementation system,'' \url{https://www.cadence.com/content/cadence-www/global/en_US/home/tools/digital-design-and-signoff/hierarchical-design-and-floorplanning/innovus-implementation-system.html}.

\bibitem{Proteus}
\BIBentryALTinterwordspacing
P.~Judd, J.~Albericio, T.~Hetherington, T.~M. Aamodt, N.~E. Jerger, and A.~Moshovos, ``Proteus: Exploiting numerical precision variability in deep neural networks,'' in \emph{Proceedings of the 2016 International Conference on Supercomputing}, ser. ICS '16.\hskip 1em plus 0.5em minus 0.4em\relax New York, NY, USA: ACM, 2016, pp. 23:1--23:12. [Online]. Available: \url{http://doi.acm.org/10.1145/2925426.2926294}
\BIBentrySTDinterwordspacing

\bibitem{EIEISCA16}
S.~Han, X.~Liu, H.~Mao, J.~Pu, A.~Pedram, M.~A. Horowitz, and W.~J. Dally, ``{EIE:} efficient inference engine on compressed deep neural network,'' in \emph{43rd {ACM/IEEE} Annual International Symposium on Computer Architecture, {ISCA} 2016, Seoul, South Korea, June 18-22, 2016}, 2016, pp. 243--254.

\bibitem{ProteusICS16}
\BIBentryALTinterwordspacing
P.~Judd, J.~Albericio, T.~Hetherington, T.~M. Aamodt, N.~E. Jerger, and A.~Moshovos, ``Proteus: Exploiting numerical precision variability in deep neural networks,'' in \emph{Proceedings of the 2016 International Conference on Supercomputing}, ser. ICS '16.\hskip 1em plus 0.5em minus 0.4em\relax New York, NY, USA: ACM, 2016, pp. 23:1--23:12. [Online]. Available: \url{http://doi.acm.org/10.1145/2925426.2926294}
\BIBentrySTDinterwordspacing

\bibitem{cacti}
HewlettPackard, ``{CACTI},'' \url{https://github.com/HewlettPackard/cacti}.

\end{thebibliography}
\bibliographystyle{IEEEtran}

\clearpage
\newpage
\appendix

\section{Our Hardware Approach}
\label{sec:hardware}
This section presents the \DPRL hardware encoder/decoder units that efficiently exploit the potential created by our quantization schemes.
Without the loss of generality, we describe compressors/decompressors that process groups of 8 FP32 values. This section assumes that the reader is aware of prior work that demonstrated why it is possible and desirable to encode tensor values using variable length containers when storing them to external DRAM, e.g.,~\cite{Deep_Compression, Proteus, EIEISCA16}. 

\noindent\textbf{Overview of Hardware: }At high-level, our hardware compressors/decompressors transparently encode/decode tensor values just before the memory controller. When values are stored to external DRAM, the encoders efficiently encode the values to use as few bits as necessary. When values are read back from external DRAM, the decoders, expand the values to the original format. This way the rest of the on-chip memory hierarchy and compute units can remain as-is.

\begin{figure}[ht]
    \centering
    \includegraphics[width=\linewidth]{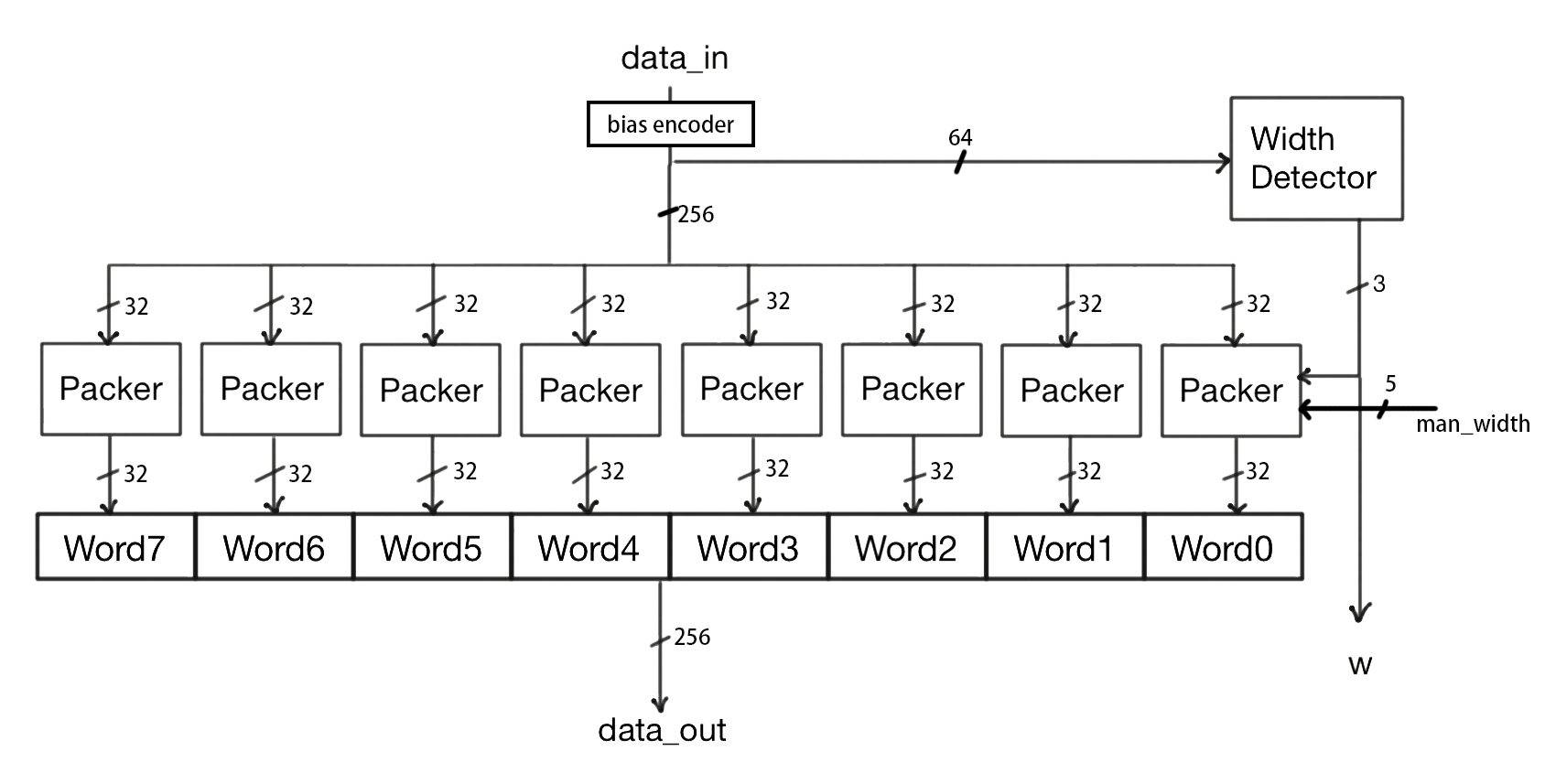}
    \caption{Compressor.}
    \label{fig:compressor}
\end{figure}

\noindent\textit{\textbf{Compressor:} }
The compressor accepts one row of $8$ numbers per cycle. In the compressor's first stage, it subtracts the fixed bias from the exponents. The resulting differences along with mantissas are then processed by $8$ Packer units and a width detector as shown in Figure~\ref{fig:compressor}. The mantissa quantizer method, whether \MP or \MC, provides the same mantissa length for all values. Each value within the row is encoded using the same number of bits, calculated as the sum of the provided mantissa bitlength and the bitlength needed to store the largest exponent difference across the row. The width detector, as its name suggests, will detect how many bits are needed to represent the exponents of the entire row. This step is accomplished by performing an OR operation on all $8$ exponent
 values, and then detecting the leading $1$. It will output a 3b number to the packer and compressor output. The exponent lengths need to be stored as metadata per row. These are stored separately, necessitating two write streams per tensor; both streams are sequential, thus DRAM-friendly. Furthermore, because there are only $3$ bits per group of $8$ values, a single access to this metadata structure yields metadata for multiple groups. Accesses to this metadata structure will thus be much less frequent than that for the value containers.
 
\begin{figure}[ht]
    \centering
    \includegraphics[width=\linewidth]{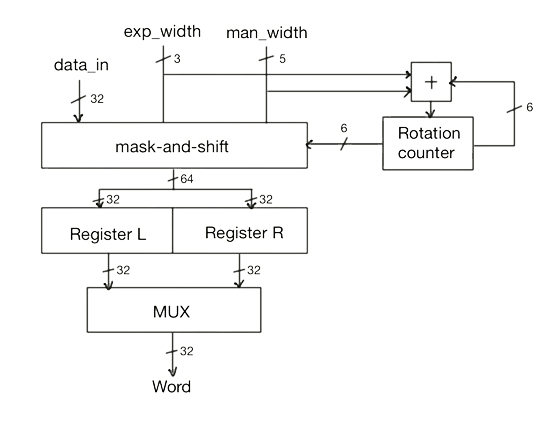}
    \caption{Packer.}
    \label{fig:packer}
\end{figure}

To avoid wide crossbars when packing/unpacking, values remain within the confines of their original format bit positions as per the method proposed in Proteus~\cite{ProteusICS16}. In contrast to Proteus, however, here every row uses a different bitlength, the values are floating-point, the bitlengths vary during runtime and per row, and we target training. Each packer, shown in Figure~\ref{fig:packer}, takes a single FP32 number masks out unused exponent and mantissa bits, and rotates the remaining bits to a position to fill in the output row. The mask is created based on the exp\_width and man\_width inputs. The rotation counter register provides the rotation count which is updated to  (exp\_width+man\_width) every cycle. The (L,R) register pair is used to tightly pack the encoded values into successive rows. They are needed since a value may now be split across two memory rows. This arrangement effectively packs the values belonging to each column tightly within a column of 32b in memory. Since each row is the same total bitlength, the $8$ packers operate in tandem filling their respective outputs at exactly the same rate. As a result, the compressor produces $8\times32$b at a time. The rate at which the outputs are produced depends on the compression rate achieved, the higher the compression, the lower the rate. 

\begin{figure}[ht]
    \centering
    \includegraphics[width=\linewidth]{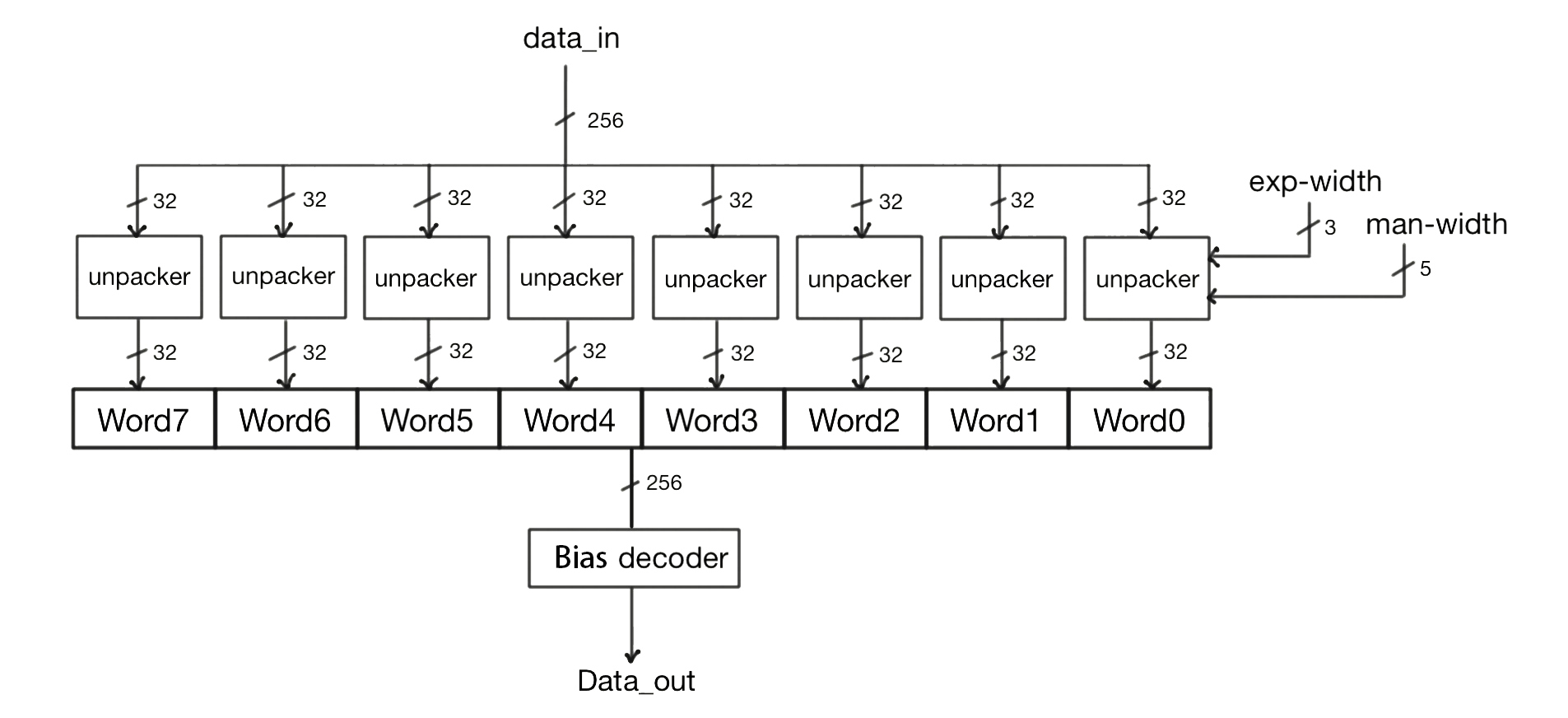}
    \caption{Decompressor.}
    \label{fig:decompressor}
\end{figure}

\noindent\textit{\textbf{Decompressor: }}
As Figure~\ref{fig:decompressor} shows, the decompressor mirrors the compressor. The inputs to the unit are a 3b exp\_width, a 5b man\_width, and a $8\times32$b compressed data input. Since the data is compressed, a single row of $8\times32$b will typically contain data from more than one original uncompressed row of FP32 numbers. The compressed data values are packed into $8$ virtual columns withing each row. Accordingly, each of the $8$ virtual columns of 32b is fed into a dedicated unpacker.

\begin{figure}[ht]
    \centering
    \includegraphics[width=0.8\linewidth]{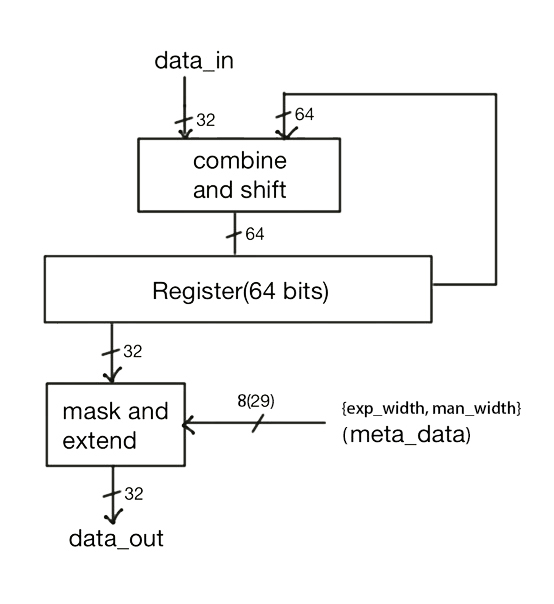}
    \caption{Unpacker.}
    \label{fig:unpacker}
\end{figure}

Each unpacker, shown in Figure~\ref{fig:unpacker}, has a wide 64b register that is internally divided into $L$ and $R$ registers of 32b. They are used in a similar fashion as the corresponding registers of the packer unit. At any point in time, one of the registers is used to accept a new row of 32b packed data whereas the other contains whichever bits from the previous row of 32b have not been used yet. The combine-and-shift will combine the input data and previous data in the register then shift to the left. The number of shifted bits is determined by the exponent and mantissa lengths of this row.
The 32b data on the left of the register are taken out and shifted to the right (zero extending the exponent). Finally, the unpacker reinserts the mantissa bits that were trimmed during compression. Since each row of data uses the same total bitlength, the unpackers operate in tandem consuming data at the same rate. The net effect is that external memory see wide accesses on both sides.

\section{Hardware Evaluation Methodology}\label{sec:eval_method}
Best practices for the evaluation of custom hardware architectures necessitate exploration and validation first via analytical modelling or via cycle-accurate simulation. Since training these networks takes several \textit{days} on actual hardware, cycle-accurate simulation of the full process is impractical. To estimate performance and energy, we use the best practice approach by analytically modelling the time and energy used per layer per pass of a baseline accelerator. To do so, we use traffic and compute counts collected during the aforementioned full training runs. We record these counts each time a layer is invoked using PyTorch hooks.  We model time and energy for memory accesses via DRAMSIM3~\cite{Dramsim3}. For modeling on-chip structures we use CACTI~\cite{cacti} for the buffers and layout measurements for the compute units and the \EC compressors/decompressors. We use a commercial 65nm process to model the processing units and \EC hardware. We implement the units in Verilog and perform synthesis via the Synopsys Design Compiler and \textit{layout} using Cadence Innovus with a target frequency of 500MHz. Synthesis uses Synopsys' commercial Building Block IP library for the target tech node. We estimate power via Innovus using traces over a representative input sample to model properly signal activity. We used nominal operating conditions to model power and latency. There are two \EC compressor/decompressor units per channel.

\begin{table}[]
\centering
\caption{Hardware Area Overhead.}
\resizebox{\columnwidth}{!}
{
\begin{tabular}{|| c | c | c | c ||}
\hline
module & area per unit ($um^{2}$) & unit number & total area ($mm^{2}$) \\
\hline
compressor & 31575.60 & 16 & 0.505 \\
\hline
decompressor & 37133.28 & 16 & 0.594 \\
\hline
accelerator & 38533.68 & 8000 & 308.27 \\
\hline
\end{tabular}
}
\label{tbl:training_hyper_parameters}
\end{table}

Due to the complexity and time cost of cycle-accurate hardware simulation, we have opted for an estimated time and energy consumption analytical model based on the proposed hardware description and the compressor-decompressor architecture. To compute the analytical model, we first analyze the network and retrieve its structure (layer input and output sizes, kernel sizes for convolutional layers, stride, bias and padding). We then calculate the compute operations that will happen for the general batch size (N) in both the forward and backward pass, as well as the number of parameters that must be stored in memory for activations, weights and gradients.

To take advantage of data reuse where possible we perform the forward pass in a layer-first order per batch. This allows us to read the weights per layer only once per batch. For the backward pass, we utilize the on-chip buffers for mini-batching with a layer-first order over a mini-batch of samples. Mini-batching reduces overall traffic by processing as many samples as possible in a layer-first order avoiding either having to spill gradients or reading and writing weights per sample per layer. The number of samples that can fit in a mini-batch depends on the layer dimensions and the size of the on-chip buffer.

Both \GMP and \GMC sample bitlengths per batch to a log file for both mantissas and exponents. These bitlengths are used to compute the number of mini-batches that can fit at every training step per layer on chip. Based on the number of sampled mini-batches (K) we compute the memory footprint generated on the forward pass for each method. After this, we calculate the footprint that stays on-chip and can be loaded from on-chip for the backward pass, and the footprint that goes to off-chip and has to be loaded to on-chip again for it. Based on these memory accesses, we use DRAMsim to simulate the number of compute-cycles that take the memory accesses to finish and we use the maximum cycles between compute and memory as the time constraint to calculate total computation time in the proposed hardware.

To calculate energy consumption and efficiency, we use the information gathered in terms of on-chip memory access cycles, off-chip memory access cycles and compute cycles. We estimate energy consumption for all components including the compressors and decompressors. We use the following equations to estimate energy consumption for our methods (all symbols are defined in Table \ref{tbl:symbols_definition_table}):

\begin{equation}
    \begin{gathered}
        {E_{\,f\!orward}} = {E_{\,compute\,fwd}} + {E_{\,o\!f\!\!f\!chip\,in\,act\,mem}} + \\ 
        {E_{\,o\!f\!\!f\!chip\,wgt\,mem}} + {E_{\,o\!f\!\!f\!chip\,out\,act\,mem}} + {E_{\,onchip\,in\,act\,mem}} + \\
        {E_{\,onchip\,wgt\,mem}} + {E_{\,onchip\,out\,act\,mem}} + {E_{\,read\,ops\,mem}} + \\
        {E_{\,decomp\,act}} + {E_{\,decomp\,wgt}} + {E_{\,comp\,act}}
    \end{gathered}
\end{equation}

\begin{equation}
    \begin{gathered}
        {E_{\,backward}} = {E_{\,compute\,bck}} + {E_{\,o\!f\!\!f\!chip\,in\,act\,mem}} + \\ 
        {E_{\,o\!f\!\!f\!chip\,wgt\,mem}} + {E_{\,onchip\,in\,act\,mem}} + \\
        {E_{\,onchip\,wgt\,mem}} + {E_{\,read\,ops\,mem}} + \\
        {E_{\,decomp\,act}} + {E_{\,decomp\,wgt}}
    \end{gathered}
\end{equation}

where,

\begin{equation}
{E_{\,o\!f\!\!f\!chip\,in\,act\,mem}} = \frac{{MemCh} \times {P_{\,DRAM}}}{Freq_{compute}} \times \\{Cycles_{\,o\!f\!\!f\!chip\,in\,act}}
\end{equation}

\begin{equation}
\begin{gathered}
{E_{\,o\!f\!\!f\!chip\,wgt\,mem}} = \frac{{MemCh} \times {P_{\,DRAM}}}{Freq_{compute}} \times \\({Cycles_{\,o\!f\!\!f\!chip\,wgt}} + {Cycles_{\,o\!f\!\!f\!chip\,wgt\,grad}})
\end{gathered}
\end{equation}

\begin{equation}
{E_{\,o\!f\!\!f\!chip\,out\,act\,mem}} = \frac{{MemCh} \times {P_{\,DRAM}}}{Freq_{compute}} \times {Cycles_{\,o\!f\!\!f\!chip\,out\,act}}
\end{equation}

\begin{equation}
{E_{\,onchip\,in\,act\,mem}} = {Cycles_{\,onchip\,in\,act\,write}} \times {P_{\,onchip\,write}}
\end{equation}

\begin{equation}
{E_{\,onchip\,wgt\,mem}} = {Cycles_{\,onchip\,wgt\,read}} \times {P_{\,onchip\,read}}
\end{equation}

\begin{equation}
    \begin{gathered}
        {E_{\,onchip\,out\,act\,mem}} = {Cycles_{\,onchip\,out\,act\,read}} \times {P_{\,onchip\,read}} + \\ {Cycles_{\,onchip\,out\,act\,write}} \times {P_{\,onchip\,write}}
    \end{gathered}
\end{equation}

\begin{equation}
{E_{\,decomp}} = {P_{\,decomp\,(comp\,ratio)}} \times \frac{Cycles_{\,comp\,to\,decomp}}{Freq_{compute}}
\end{equation}
\begin{equation}
{E_{\,comp}} = {P_{\,comp\,(comp\,ratio)}} \times \frac{Cycles_{\,decomp\,to\,comp}}{Freq_{compute}}
\end{equation}

\begin{equation}
{E_{\,decomp\,act}} = {E_{\,decomp\,act(comp\,ratio)}}
\end{equation}

\begin{equation}
{E_{\,decomp\,wgt}} = {E_{\,decomp\,wgt(comp\,ratio)}}
\end{equation}

\begin{equation}
{E_{\,comp\,act}} = {E_{\,comp\,act(comp\,ratio)}}
\end{equation}

\begin{table}[H]
\centering
\caption{$P()$ terms: Power consumption as a function compression ratio. }
\resizebox{\columnwidth}{!}
{
\begin{tabular}{|| c | c | c ||}
\hline
Compression ratio & Compressor power (mW) & Decompressor power (mW) \\
\hline
0.143 - 0.263 & 10.87 & 13.84 \\
\hline
0.264 - 0.388 & 12.18 & 14.72 \\
\hline
0.389 - 0.513 & 12.65 & 15.97 \\
\hline
0.514 - 0.638 & 13.44 & 15.76 \\
\hline
0.639 - 0.763 & 14.98 & 15.42 \\
\hline
\end{tabular}

\label{tbl:comp_decomp_power_consumptions}
}
\end{table}

\clearpage

\begin{table*}[hbtp]
\centering
\caption{Symbols definition table.}
\begin{adjustbox}{width=1\textwidth}
\begin{tabular}{|| c | c ||}
\hline
Symbol & Definition \\
\hline
${E_{\,compute\,f\!wd}}$ & \begin{tabular}{@{}c@{}} Energy consumption of the compute module for\\ the entirety of the computations in the forward pass \end{tabular} \\
\hline
${E_{\,compute\,bck}}$ & \begin{tabular}{@{}c@{}} Energy consumption of the compute module for\\ the entirety of the computations in the backward pass \end{tabular} \\
\hline
${E_{\,o\!f\!\!f\!chip\,in\,act\,mem}}$ & \begin{tabular}{@{}c@{}} Energy consumption of the offchip memory transfers\\ for the network input activations \end{tabular} \\
\hline
${E_{\,o\!f\!\!f\!chip\,wgt\,mem}}$ & \begin{tabular}{@{}c@{}} Energy consumption of the offchip memory transfers\\ for the network weights \end{tabular} \\
\hline
${E_{\,o\!f\!\!f\!chip\,out\,act\,mem}}$ & \begin{tabular}{@{}c@{}} Energy consumption of the offchip memory transfers\\ for the network output activations \end{tabular} \\
\hline
${E_{\,onchip\,in\,act\,mem}}$ & \begin{tabular}{@{}c@{}} Energy consumption of the onchip memory transfers\\ for the network input activations \end{tabular} \\
\hline
${E_{\,onchip\,wgt\,mem}}$ & \begin{tabular}{@{}c@{}} Energy consumption of the onchip memory transfers\\ for the network weights \end{tabular} \\
\hline
${E_{\,onchip\,out\,act\,mem}}$ & \begin{tabular}{@{}c@{}} Energy consumption of the onchip memory transfers\\ for the network output activations \end{tabular} \\
\hline
${E_{\,read\,ops\,mem}}$ & \begin{tabular}{@{}c@{}} Energy consumption of loading \\ operations from memory \end{tabular} \\
\hline
${E_{\,decomp\,act}}$ & \begin{tabular}{@{}c@{}} Energy consumption of decompressing \\ activations in the decompressor \end{tabular} \\
\hline
${E_{\,decomp\,wgt}}$ & \begin{tabular}{@{}c@{}} Energy consumption of decompressing \\ weights in the decompressor \end{tabular} \\
\hline
${E_{\,comp\,act}}$ & \begin{tabular}{@{}c@{}} Energy consumption of compressing \\ activations in the compressor \end{tabular} \\
\hline
${P_{\,decomp\,(comp\,ratio)}}$ & \begin{tabular}{@{}c@{}} Power consumption by the decompressor when loading data\\ from offchip memory at a specific compression ratio (see Table \ref{tbl:comp_decomp_power_consumptions})\end{tabular}  \\
\hline
${P_{\,comp\,(comp\,ratio)}}$ & \begin{tabular}{@{}c@{}} Power consumption by the compressor when writing data\\ to offchip memory at a specific compression ratio (see Table \ref{tbl:comp_decomp_power_consumptions}) \end{tabular} \\
\hline
$MemCh$ & Number of available memory channels \\
\hline
$P_{DRAM}$ & Power consumption of offchip DRAM \\
\hline
${Freq_{compute}}$ & Clock frequency of the hardware accelerator \\
\hline
${Cycles_{\,o\!f\!\!f\!chip\,in\,act}}$ & Compute cycles taken to read input activations from offchip memory \\
\hline
${Cycles_{\,o\!f\!\!f\!chip\,wgt}}$ & Compute cycles taken to read weights from offchip memory \\
\hline
${Cycles_{\,o\!f\!\!f\!chip\,wgt\,grad}}$ & Compute cycles taken to read weight gradients from offchip memory \\
\hline
${Cycles_{\,o\!f\!\!f\!chip\,out\,act}}$ & Compute cycles taken to read output activations from offchip memory \\
\hline
${Cycles_{\,onchip\,in\,act\,write}}$ & Compute cycles taken to read input activations from onchip memory \\
\hline
${Cycles_{\,onchip\,wgt\,read}}$ & Compute cycles taken to read weights from onchip memory \\
\hline
${Cycles_{\,onchip\,out\,act\,read}}$ & Compute cycles taken to read output activations from onchip memory \\
\hline
${Cycles_{\,onchip\,out\,act\,write}}$ & Compute cycles taken to write output activations to onchip memory \\
\hline
${P_{\,onchip\,write}}$ & Power consumption of a word write to onchip memory \\
\hline
${P_{\,onchip\,read}}$ & Power consumption of a word read from onchip memory \\
\hline
${Cycles_{\,comp\,to\,decomp}}$ & Compute cycles taken to decompress compressed data \\
\hline
${Cycles_{\,decomp\,to\,comp}}$ & Compute cycles taken to compress data \\
\hline
\end{tabular}
\end{adjustbox}

\label{tbl:symbols_definition_table}
\end{table*}

\end{document}